%% file: main.tex
\documentclass[10pt,twocolumn,letterpaper]{article}

\usepackage[pagenumbers]{cvpr} 

\input{preamble}
\newtoggle{arXiv}

\definecolor{cvprblue}{rgb}{0.21,0.49,0.74}
\usepackage[pagebackref,breaklinks,colorlinks,citecolor=cvprblue]{hyperref}
\toggletrue{arXiv}

\title{3DGS-Avatar: Animatable Avatars via Deformable 3D Gaussian Splatting}

\author{
Zhiyin Qian\textsuperscript{1} \quad 
Shaofei Wang\textsuperscript{1,2,3} \quad
Marko Mihajlovic\textsuperscript{1} \quad
Andreas Geiger\textsuperscript{2,3} \quad
Siyu Tang\textsuperscript{1}\\
\textsuperscript{1}ETH Z\"{u}rich \quad 
\textsuperscript{2}University of T\"{u}bingen \quad 
\textsuperscript{3}T\"{u}bingen AI Center
}

\begin{document}
\twocolumn[{%
    \renewcommand\twocolumn[1][]{#1}%
    \setlength{\tabcolsep}{0.0mm} 

    \newcommand{\sz}{0.125}  

    \maketitle
    \begin{center}
        \newcommand{\teaserwidth}{\textwidth}
        \includegraphics[width=\linewidth]{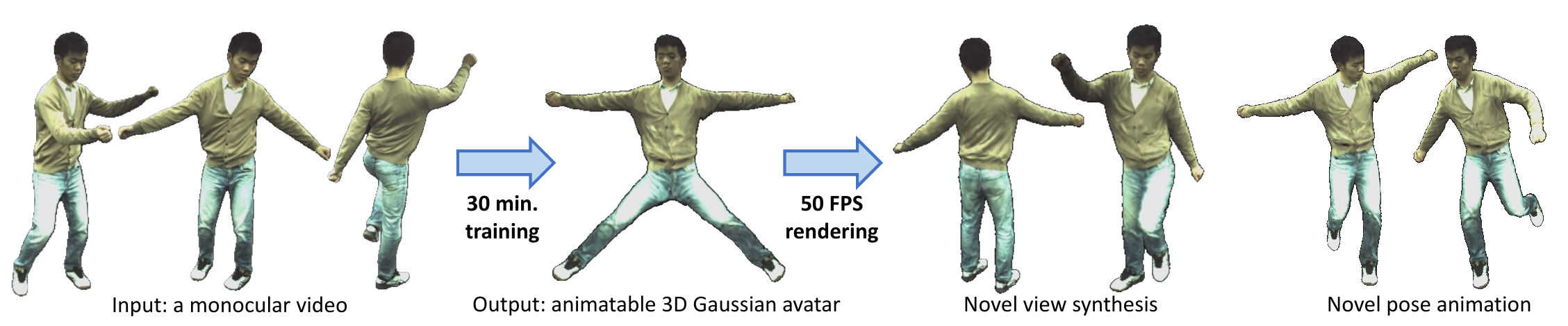}
    \captionof{figure}{
    \textbf{3DGS-Avatar.} We develop an efficient method for creating animatable avatars from monocular videos, leveraging 3D Gaussian Splatting~\cite{kerbl3Dgaussians}. Given a short sequence of dynamic human with a tracked skeleton and foreground masks, our method creates an avatar within \textbf{30 minutes} on a single GPU, supports animation and novel view synthesis at over \textbf{50 FPS}, and achieves comparable or better rendering quality to the state-of-the-art~\cite{weng2022humannerf,ARAH:ECCV:2022} that requires over 8 GPU days to train, takes several seconds to render a single image, and relies on pre-training on clothed human scans~\cite{ARAH:ECCV:2022}.
    } 

    \label{fig:teaser}
    \end{center}%
}]

\input{sec/0_abstract}    
\input{sec/1_intro}
\input{sec/2_related_works}
\input{sec/3_preliminary}

\input{sec/4_methods}
\input{sec/5_experiments}
\input{sec/6_conclusion}

{
    \small
    \bibliographystyle{ieeenat_fullname}
    \bibliography{main}
}

\input{sec/X_suppl}

\end{document}

%% file: preamble.tex
\usepackage[dvipsnames]{xcolor}

\usepackage{epsfig}
\usepackage{xcolor}
\usepackage{graphicx}
\usepackage{float}
\usepackage[utf8]{inputenc}

\usepackage{caption}
\usepackage{subcaption}
\usepackage{array}
\usepackage{algorithm}
\usepackage{algpseudocode}
\usepackage{multirow}
\usepackage{etoolbox}
\usepackage{anyfontsize}
\usepackage{booktabs}
\usepackage{colortbl}
\usepackage{bm}
\usepackage{pifont}
\usepackage{adjustbox}
\usepackage[accsupp]{axessibility}

\newcommand{\myparagraph}[1]{\noindent\textbf{#1}}

\newcommand{\cmark}{\textcolor{green}{\ding{51}}} 
\newcommand{\ccmark}{\cmark\kern-0.4em\cmark} 
\newcommand{\xmark}{\textcolor{red}{\ding{55}}} 

\newcolumntype{R}[2]{%
    >{\adjustbox{angle=#1,lap=\width-(#2)}\bgroup}%
    l%
    <{\egroup}%
}
\newcommand*\rot{\multicolumn{1}{R{35}{1em}}}

\newcommand{\colorfirst}{255, 153, 153}
\newcommand{\colorsecond}{255, 204, 153}

%% file: sec/0_abstract.tex
\begin{abstract}
We introduce an approach that creates animatable human avatars from monocular videos using 3D Gaussian Splatting (3DGS). Existing methods based on neural radiance fields (NeRFs) achieve high-quality novel-view/novel-pose image synthesis but often require days of training, and are extremely slow at inference time.
Recently, the community has explored fast grid structures for efficient training of clothed avatars. 
Albeit being extremely fast at training, these methods can barely achieve an interactive rendering frame rate with around 15 FPS. 
In this paper, we use 3D Gaussian Splatting and learn a non-rigid deformation network to reconstruct animatable clothed human avatars that can be trained within 30 minutes and rendered at real-time frame rates (50+ FPS).
Given the explicit nature of our representation, we further introduce as-isometric-as-possible regularizations on both the Gaussian mean vectors and the covariance matrices, enhancing the generalization of our model on highly articulated unseen poses. Experimental results show that our method achieves comparable and even better performance compared to state-of-the-art approaches on animatable avatar creation from a monocular input, while being 400x and 250x faster in training and inference, respectively. Please see our project page at \href{https://neuralbodies.github.io/3DGS-Avatar}{https://neuralbodies.github.io/3DGS-Avatar}.
\end{abstract}

%% file: sec/1_intro.tex
\section{Introduction}
\label{sec:intro}

Reconstructing clothed human avatars from image inputs presents a significant challenge in computer vision, yet holds immense importance due to its applications in virtual reality, gaming, and e-commerce. 
Traditional methods often rely on dense, synchronized multi-view inputs, which may not be readily available in more practical scenarios. Recent advances in implicit neural fields~\cite{mildenhall2020nerf,sitzmann2021lfns,suhail2022lightfield,suhail2022generalizable,wang2022r2l,Niemeyer2020CVPR,yariv2020multiview,volsdf:NeurIPS:2021,wang2021neus,Oechsle2021ICCV,Peng2021SAP} have enabled
high-quality reconstruction of geometry~\cite{HNeRF:NeurIPS:2021,peng2022animatable,ARAH:ECCV:2022,guo2023vid2avatar} and appearance~\cite{peng2020neural,peng2021animatable,NARF:ICCV:2021,liu2021neural,raj2020anr,li2022tava,jiang2022neuman,weng2022humannerf,yu2023monohuman} of clothed human bodies from
sparse multi-view or monocular videos. 
Animation of such reconstructed clothed human bodies
is also possible by learning the geometry and appearance representations in a
predefined canonical pose~\cite{weng2022humannerf,ARAH:ECCV:2022,peng2021animatable,li2022tava,jiang2022neuman,yu2023monohuman}.

To achieve state-of-the-art rendering quality, existing methods rely on training
a neural radiance field (NeRF)~\cite{mildenhall2020nerf} combined with either
explicit body articulation~\cite{jiang2022neuman,peng2021animatable,ARAH:ECCV:2022,guo2023vid2avatar,li2022tava,peng2022animatable,jiang2022instantavatar,weng2022humannerf,yu2023monohuman} or conditioning the NeRF on human body related
encodings~\cite{HNeRF:NeurIPS:2021,NARF:ICCV:2021,ANeRF:NeurIPS:2021,peng2020neural}. They often employ large multi-layer perceptrons (MLPs) to model the neural radiance field, which are computationally demanding, leading to prolonged
training (days) and inference (seconds) time.
This computational expense poses a significant challenge for practical applications of these state-of-the-art methods in real-time applications.

With recent advances in efficient learning of implicit neural fields, training
time of NeRFs has been reduced to
minutes~\cite{yu_and_fridovichkeil2021plenoxels,NeRFAcc:ICCV:2023,mueller2022instant,TensoRF:ECCV:2022,
DVGO:CVPR:2022}. There are also works targeting fast inference of
pretrained NeRFs~\cite{yu2021plenoctrees,yariv2023bakedsdf,Reiser2023SIGGRAPH}. 
Inspired by these developments, several avatar reconstruction methods have been
tailored to fast training ~\cite{instant_nvr,jiang2022instantavatar} or fast
inference~\cite{kwon2023deliffas,chen2023uv,peng2023mlpmaps}.
However, to the best of our knowledge, there currently exists no method that simultaneously achieves both fast training and {\it real-time} inference for {\it animatable} avatar reconstruction from just monocular videos.

Point-based rendering~\cite{xu2022point,Zheng2023pointavatar,ruckert2021adop,zhang2022differentiable,su2023npc, zheng2022structured,SMPLpix:WACV:2020} has emerged as an efficient alternative to NeRFs for fast inference. With the recently proposed 3D Gaussian
Splatting (3DGS)~\cite{kerbl3Dgaussians}, it is possible to achieve state-of-the-art rendering quality using only a fraction of NeRFs' inference time and comparatively fast training for static scene reconstruction.

Leveraging the capabilities of 3DGS, we demonstrate its application in modeling animatable clothed avatars using monocular videos. 
Our approach effectively integrates rigid human articulation with a non-rigid deformation field within the 3DGS framework. We use a small multi-layer perceptron (MLP) to decode color. This MLP is designed to be responsive to local non-rigid deformations and dynamic lighting conditions, ensuring a more realistic and responsive rendering of the avatar's appearance. Furthermore, we apply as-isometric-as-possible regularizations~\cite{kmp_shape_space_sig_07,Prokudin_2023_ICCV} to \textit{both} the Gaussian mean vectors and the covariance matrices, which helps maintain the geometric consistency and realistic deformation of the avatar, particularly in dynamic and varied poses. 

Our experimental results show that our method is comparable to or better than current state-of-the-art \cite{ARAH:ECCV:2022,weng2022humannerf} in animatable avatar creation from monocular inputs, achieving training speed 400 times faster and inference speed 250 times quicker.  
Compared to methods that focus on fast training~\cite{instant_nvr,jiang2022instantavatar}, our method, despite being slower in training, can model pose-dependent non-rigid deformation and produce significantly better rendering quality, while being 3 times faster in terms of rendering.
We provide an overview of the comparison to major prior works in \cref{tab:comparison}.
In summary, our work makes the following contributions: 
\begin{itemize}
    \item We introduce 3D Gaussian Splatting to animatable human avatars reconstruction from monocular videos. 
    \item We develop a simple yet effective deformation network as well as regularization terms that effectively drive 3D Gaussian Splats to handle highly articulated and out-of-distribution poses. 
    \item Our method is the first, to our knowledge, to simultaneously deliver high-quality rendering, model pose-dependent non-rigid deformation, generalize effectively to unseen poses, and achieve fast training (less than 30 minutes) and real-time rendering speed (50+ FPS).
\end{itemize}

\begin{table}
    \begin{center}
    \small
    \begin{tabular}{ccccc|l}
    \rot{pose-dependent deformation} 
    & \rot{novel pose animation}  
    & \rot{fast training} 
    & \rot{real-time rendering}  
    & \rot{monocular input} &  \\
    \hline\hline
    \xmark
    & \xmark
    & \xmark
    & \xmark
    & \cmark 
    & NeuralBody~\cite{peng2020neural}
    \\
    \cmark
    & \xmark
    & \xmark
    & \xmark
    & \cmark
    & HumanNeRF~\cite{weng2022humannerf}
    \\
    \cmark
    & \cmark
    & \xmark
    & \xmark
    & \cmark
    & ARAH~\cite{ARAH:ECCV:2022}
    \\
    \cmark
    & \xmark
    & \ccmark
    & \xmark
    & \cmark
    & Instant-NVR~\cite{instant_nvr}
    \\
    \xmark
    & \cmark
    & \ccmark
    & \xmark
    & \cmark
    & InstantAvatar~\cite{jiang2022instantavatar}
    \\
    \cmark
    & \cmark
    & \xmark
    & \xmark
    & \cmark
    & MonoHuman~\cite{yu2023monohuman}
    \\
    \cmark
    & \cmark
    & \xmark
    & \xmark
    & \xmark
    & UV-Volumes~\cite{chen2023uv}
    \\
    \cmark
    & \cmark
    & \xmark
    & \cmark
    & \xmark
    & DELIFFAS~\cite{kwon2023deliffas}
    \\
    \hline
    \cmark
    & \cmark
    & \cmark
    & \cmark
    & \cmark
    & \textbf{3DGS-Avatar (Ours)}
    \end{tabular}
    \end{center}
    \vspace{-3mm}
    \caption{\textbf{Comparison to SoTA.}
    Instant-NVR~\cite{instant_nvr} and InstantAvatar~\cite{jiang2022instantavatar} achieve instant training within $5$ minutes.
    For real-time rendering, we require a frame rate over 30 FPS. Note that while UV-Volumes~\cite{chen2023uv} claims real-time freeview rendering, they only achieve $14$ FPS on novel pose synthesis due to the slow generation of their UV Volume. }
    \label{tab:comparison}
\end{table}

%% file: sec/2_related_works.tex
\section{Related Works}
\label{sec:related}

\myparagraph{Neural rendering for clothed human avatars.}
Since the seminal work of Neural Radiance Fields (NeRF)~\cite{mildenhall2020nerf},
there has been a surge of research on neural rendering for clothed human avatars.
The majority of the works focus on either learning a NeRF conditioned on human 
body related encodings~\cite{HNeRF:NeurIPS:2021,NARF:ICCV:2021,ANeRF:NeurIPS:2021}, or learning a canonical NeRF representation and
warp camera rays from the observation space to the canonical space to 
query radiance and density values from the canonical NeRF~\cite{jiang2022neuman,peng2021animatable,ARAH:ECCV:2022,guo2023vid2avatar,li2022tava,peng2022animatable,jiang2022instantavatar,weng2022humannerf,yu2023monohuman}. Most of these
works rely on large multi-layer perceptrons (MLPs) to model the underlying neural
radiance field, which are computationally expensive, resulting in prolonged training
(days) and inference (seconds) time.

With recent advances in accelerated data structures for neural fields, there has
been several works targeting fast inference and fast training of NeRFs for
clothed humans.~\cite{jiang2022instantavatar} proposes to use iNGP~\cite{mueller2022instant} as the underlying
representation for articulated NeRFs, which enables fast training (less than 5
minutes) and interactive rendering speed (15 FPS) but ignores pose-dependent non-rigid deformations.~\cite{instant_nvr} also
utilizes iNGP and represents non-rigid deformations in the UV space, which enables
fast training and modeling of pose-dependent non-rigid deformations. However, as
we will show in our experiments,~\cite{instant_nvr}'s parametrization of
non-rigid deformations result in blurry renderings.~\cite{chen2023uv} proposes
to generate a pose-dependent UV volume for efficient free-view synthesis.
However, their UV-volume generation process is slow (20 FPS), making novel pose
synthesis less efficient (only 14 FPS).~\cite{kwon2023deliffas} also employs UV-based
rendering to achieve real-time rendering of dynamic clothed humans, but only
works on dense multi-view inputs. Extending~\cite{yu2021plenoctrees},
\cite{FourierOctree:CVPR:2022,NSR:SIGASIA:2022} applied Fourier transform for
compressing human performance capture data, albeit with limitations on dense
multi-view data (60-80 views) and non-generalizability of the Fourier basis
representation to unseen poses beyond the training dataset.
In contrast to all these works, our method achieves state-of-the-art rendering quality and speed with less than 30 minutes
of training time from a single monocular video input.

\myparagraph{Dynamic 3D gaussians.}
Point-based rendering~\cite{xu2022point,Zheng2023pointavatar, ruckert2021adop,zhang2022differentiable,su2023npc, zheng2022structured,SMPLpix:WACV:2020} has also been shown to be an efficient alternative to NeRFs for fast inference and training. Extending point cloud to 3D Gaussians, 3D Gaussian Splatting (3DGS)~\cite{kerbl3Dgaussians} models the rendering process as splatting a set of 3D Gaussians onto image plane via alpha blending, achieving state-of-the-art rendering quality with real-time inference speed and fast training given multi-view inputs. 

Given the great performance on both quality and speed of 3DGS, a rich set of works has further explored the 3D Gaussian representation for dynamic scene reconstruction. \cite{kerbl3Dgaussians} proposed to optimize the position and shape of each 3D Gaussian on a frame-by-frame basis and simultaneously performed 6-DOF dense tracking for free. Their model size, however, increases with the temporal dimension. \cite{yang2023deformable3dgs, wu20234dgaussians} maintain a single set of 3D Gaussians in a canonical space and deform them to each frame via learning a time-dependent deformation field, producing state-of-the-art results in terms of both rendering quality and speed. \cite{yang2023gs4d} augments 3D Gaussians with temporal dimension into 4D Gaussian primitives to approximate the underlying spatiotemporal 4D volume of the dynamic scene. While such methods show promising results, they are only applicable to either synthetic datasets with fast camera movement and slow object motion or forward-facing real scenes with limited object movements, thus unable to handle the immense displacement of the articulated human body. To address this problem, our approach utilizes a statistical human body model~\cite{SMPL:2015} for articulation and applies regularization to reduce the overfitting of the deformation field.

\myparagraph{Concurrent works.}
Concurrent with our method, many recent works also seek to combine 3DGS with human articulation prior for avatar reconstruction. We provide a comparison of our approach to concurrent works in \cref{tab:comparison_concurrent}. D3GA~\cite{Zielonka2023Drivable3D} proposed to embed 3D Gaussians in tetrahedral cages and utilize cage deformations for drivable avatar animation. However, they use dense calibrated multi-view videos as input and require an additional 3D scan to generate the tetrahedral mesh template. Li~\etal~\cite{li2023animatable} focused on generating avatars with a detailed appearance from multi-view videos by post-processing radiance field renderings with 2D CNNs, which limits their rendering speed. Along with~\cite{jena2023splatarmor,moreau2023human}, these works fail to achieve fast training with relatively complex pipelines. Similar to our approach, Ye~\etal~\cite{ye2023animatable} deforms 3D Gaussians in canonical space via pose-dependent deformation and rigid articulation, but they still require 2 hours for training and do not show results on monocular inputs. HUGS~\cite{kocabas2023hugs} learns a background model along with the animatable human avatar, but they fail to take pose-dependent cloth deformation into account. Several other works~\cite{lei2023gart,liu2023animatable,hu2023gauhuman} also neglect pose-dependent cloth deformation to achieve even faster training (in 5 minutes) and rendering (150+ FPS). We argue that our method strikes a good balance between quality and speed compared to concurrent works, as being the only method simultaneously achieving the properties listed in \cref{tab:comparison_concurrent}.

\begin{table}
    \begin{center}
    \small
    \begin{tabular}{ccccc|l}
    \rot{pose-dependent deformation} 
    & \rot{novel pose animation}  
    & \rot{fast training} 
    & \rot{real-time rendering}  
    & \rot{monocular input} &  \\
    \hline\hline
    \cmark
    & \cmark
    & \xmark
    & \cmark
    & \xmark
    & D3GA~\cite{Zielonka2023Drivable3D}
    \\
    \cmark
    & \cmark
    & \xmark
    & \xmark
    & \xmark
    & Li~\etal~\cite{li2023animatable}
    \\
    \cmark
    & \cmark
    & \xmark
    & \cmark
    & \cmark
    & SplatArmor~\cite{jena2023splatarmor}
    \\
    \cmark
    & \cmark
    & \xmark
    & \xmark
    & \xmark
    & Moreau~\etal~\cite{moreau2023human}
    \\
    \cmark
    & \cmark
    & \cmark
    & \cmark
    & \xmark
    & Ye~\etal~\cite{ye2023animatable}
    \\
    \xmark
    & \cmark
    & \cmark
    & \cmark
    & \cmark
    & HUGS~\cite{kocabas2023hugs}
    \\
    \xmark
    & \cmark
    & \ccmark
    & \ccmark
    & \cmark
    & GART~\cite{lei2023gart}
    \\
    \xmark
    & \cmark
    & \ccmark
    & \ccmark
    & \cmark
    & Liu~\etal~\cite{liu2023animatable}
    \\
    \xmark
    & \xmark
    & \ccmark
    & \ccmark
    & \cmark
    & GauHuman~\cite{hu2023gauhuman}
    \\
    \hline
    \cmark
    & \cmark
    & \cmark
    & \cmark
    & \cmark
    & \textbf{3DGS-Avatar (Ours)}
    \end{tabular}
    \end{center}
    \vspace{-3mm}
    \caption{\textbf{Comparison to Concurrent Works.}
     }
    \label{tab:comparison_concurrent}
\end{table}

%% file: sec/3_preliminary.tex
\section{Preliminaries}
\label{sec:prelim}

\myparagraph{Linear Blend Skinning.}
To model human articulations, a widely adopted paradigm is to represent geometry and appearance in a shared canonical space~\cite{jiang2022neuman,peng2021animatable,ARAH:ECCV:2022,guo2023vid2avatar,li2022tava,peng2022animatable,jiang2022instantavatar,weng2022humannerf} and use Linear Blend Skinning (LBS)~\cite{SCAPE,Hasler2009CGF,SMPL:2015,STAR:ECCV:2020,Pavlakos_2018_CVPR,Xu_2020_CVPR} to deform the parametric human body under arbitrary poses. Given a point $\mathbf{x}_c$ in canonical space, the LBS function takes a set of rigid bone transformations $\{\mathbf{B}_b\}_{b=1}^B$ and computes its correspondence $\mathbf{x}_o$ in the observation space: 
%
\begin{equation}
    \mathbf{x_o}=LBS_{\sigma_w}(\mathbf{x}_c;\{\mathbf{B}_b\})
\end{equation}
Assuming an underlying SMPL model, we use a total of $B=24$ bone transformations, each represented by a $4 \times 4$ rotation-translation matrix, which are then linearly blended via a set of skinning weights $\mathbf{w}\in [0,1]^B,s.t.\sum_{b=1}^B\mathbf{w}_b=1$, modeled by a coordinate-based neural skinning field $f_{\sigma_w}(\mathbf{x}_c)$~\cite{SCANimate:CVPR:21,LEAP:CVPR:21,Chen2023PAMI,MetaAvatar:NeurIPS:2021,Chen2021ICCV}. The forward linear blend skinning function can thus be formulated as:
\begin{equation}
    \mathbf{x}_o=LBS_{\sigma_w}(\mathbf{x}_c;\{\mathbf{B}_b\})
    =\sum\nolimits_{b=1}^B f_{\sigma_w}(\mathbf{x}_c)_b\mathbf{B}_b\mathbf{x}_c
    \label{eq:lbs}
\end{equation}
Compared to prior works that search canonical correspondences of points in observation space~\cite{weng2022humannerf,ARAH:ECCV:2022,jiang2022instantavatar}, our method requires no inverse skinning which is typically difficult to compute and often leads to multiple solutions~\cite{Chen2023PAMI,Chen2021ICCV}. A similar technique has been employed in~\cite{Zheng2023pointavatar} for face avatar modeling.

\myparagraph{3D Gaussian Splatting.}
3DGS~\cite{kerbl3Dgaussians} utilizes a set of 3D Gaussian primitives $\{\mathcal{G}\}$ as static scene representation which can be rendered in real-time via differentiable rasterization. Each 3D Gaussian $\mathcal{G}$ is defined by its mean $\mathbf{x}$, covariance $\bm{\Sigma}$, opacity $\alpha$ and view-dependent color represented by spherical harmonics coefficients $\mathbf{f}$. To ensure positive semi-definiteness, the covariance matrix is represented by a scaling matrix $\mathbf{S}$ and rotation matrix $\mathbf{R}$.
%
%
In practice, we store the diagonal vector $\mathbf{s}\in\mathbb{R}^3$ of the scaling matrix and a quaternion vector $\mathbf{q}\in\mathbb{R}^4$ to represent rotation, which can be trivially converted to a valid covariance matrix.

The 3D Gaussians are projected to the 2D image plane during the rendering process and accumulated via alpha blending. Given a viewing transformation $\mathbf{W}$ and the Jacobian of the affine approximation of the projective transformation $\mathbf{J}$, the 2D covariance matrix in camera coordinate~\cite{zwicker2001ewa} is given by $\bm{\Sigma}'=\left(\mathbf{J}\mathbf{W}\bm{\Sigma} \mathbf{W}^T\mathbf{J}^T\right)_{1:2,1:2}$.
%
%
The pixel color $C$ is thus computed by blending 3D Gaussian splats that overlap at the given pixel, sorted according to their depth:
\begin{equation}
    C=\sum\nolimits_i\left(\alpha'_i\prod\nolimits_{j=1}^{i-1}(1-\alpha'_j)\right)c_i
    \label{eq:render}
\end{equation}
where $\alpha'_i$ denotes the learned opacity $\alpha_i$ weighted by the probability density of $i$-th projected 2D Gaussian at the target pixel location. $c$ denotes the view-dependent color computed from stored SH coefficients $\mathbf{f}$.

The 3D Gaussians $\{\mathcal{G}\}$ are optimized via a photometric loss. During optimization, 3DGS adaptively controls the number of 3D Gaussians via periodic densification and pruning, achieving self-adaptive convergence to an optimal density distribution of 3D Gaussians that well represents the scene.

%% file: sec/4_methods.tex
\begin{figure*}
\centering
\includegraphics[width=1.\textwidth]{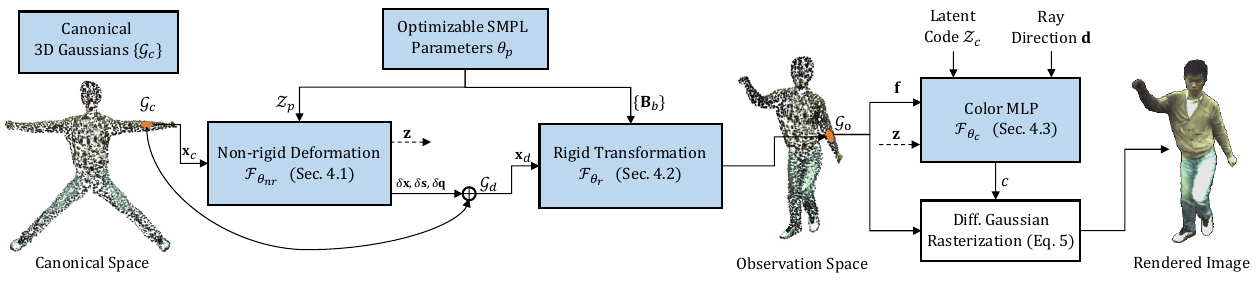}
\caption{\textbf{Our framework for creating animatable avatars from monocular videos}. We first initialize a set of 3D Gaussians in the canonical space via sampling points from a SMPL mesh. Each canonical Gaussian $\mathcal{G}_c$ goes through a non-rigid deformation module $\mathcal{F}_{\theta_{nr}}$ conditioned on an encoded pose vector $\mathcal{Z}_p$ (\cref{sec:non_rigid}) to account for pose-dependent non-rigid cloth deformation. This module outputs a non-rigidly deformed 3D Gaussian $\mathcal{G}_d$ and a pose-dependent latent feature $\mathbf{z}$. The non-rigidly deformed 3D Gaussian $\mathcal{G}_d$ is  transformed to the observation space $\mathcal{G}_o$ (\cref{sec:rigid}) via LBS with learned neural skinning $\mathcal{F}_{\theta_r}$. The Gaussian feature $\mathbf{f}$, the pose-dependent feature $\mathbf{z}$, a per-frame latent code $\mathcal{Z}_c$, and the ray direction $\mathbf{d}$ are propagated through a small MLP $\mathcal{F}_{\theta_c}$ to decode the view-dependent color $c$ for each 3D Gaussian. Finally, the observation space 3D Gaussians $\{\mathcal{G}_o\}$ and their respective color values are accumulated via differentiable Gaussian rasterization (\cref{eq:render}) to render the image. }
\label{fig:overview}
\end{figure*}
\section{Methods}
\label{sec:methods}
We illustrate the pipeline of our method in \cref{fig:overview}. The input to our method is a monocular video with a calibrated camera, fitted SMPL parameters, and foreground masks. Our method optimizes a set of 3D Gaussians in canonical space, which is then deformed to the observation space and rendered from the given camera. For a set of 3D Gaussians $\{\mathcal{G}^{(i)}\}_{i=1}^N$, we store the following properties at each point: position $\mathbf{x}$, scaling factor $\mathbf{s}$, rotation quaternion $\mathbf{q}$, opacity $\alpha$ and a color feature vector $\mathbf{f}$.
We start by randomly sampling $N=50k$ points on the canonical SMPL~\cite{SMPL:2015} mesh surface as initialization of canonical 3D Gaussians $\{\mathcal{G}_c\}$. Inspired by HumanNeRF~\cite{weng2022humannerf}, we decompose the complex human deformation into a non-rigid part that encodes pose-dependent cloth deformation, and a rigid transformation controlled by the human skeleton.

\subsection{Pose-dependent Non-rigid Deformation}
\label{sec:non_rigid}
We formulate the non-rigid deformation module as:
\begin{equation}
    \{\mathcal{G}_d\}=\mathcal{F}_{\theta_{nr}}\left(\{\mathcal{G}_c\}; \mathcal{Z}_p\right)
    \label{eq:nr_gaussian_set}
\end{equation}
where $\{\mathcal{G}_d\}$ represents the non-rigidly deformed 3D Gaussians. $\theta_{nr}$ represents the learnable parameters of the non-rigid deformation module. $\mathcal{Z}_p$ is a latent code which encodes SMPL pose and shape $(\theta,\beta)$ using a lightweight hierarchical pose encoder~\cite{LEAP:CVPR:21}. Specifically, the deformation network $f_{\theta_{nr}}$ takes the canonical position $\mathbf{x}_c$, the pose latent code $\mathcal{Z}_p$ as inputs and outputs the offsets of the Gaussian's position, scale, rotation, along with a feature vector $\mathbf{z}$:
%
\begin{equation}
    (\delta\mathbf{x},\delta\mathbf{s},\delta\mathbf{q},\mathbf{z})
    =f_{\theta_{nr}}\left(\mathbf{x}_c;\mathcal{Z}_p\right)
    \label{eq:nr_gaussian}
\end{equation}
We use a multi-level hash grid~\cite{mueller2022instant} to encode 3D positions as spatial features, which are then concatenated with the pose latent code $\mathcal{Z}_p$ and fed into a shallow MLP with 2 hidden layers and a width of 128. The canonical Gaussian is deformed by:
\begin{align}
    \mathbf{x}_d&=\mathbf{x}_c+\delta\mathbf{x} \label{eq:nr_deform_xyz}\\
    \mathbf{s}_d&=\mathbf{s}_c\cdot\exp(\delta\mathbf{s}) \label{eq:nr_deform_scale}\\
    \mathbf{q}_d&=\mathbf{q}_c\cdot[1,\delta q_1,\delta q_2, \delta q_3] \label{eq:nr_deform_rot}
\end{align}
note that the $\cdot$ operator on quaternions is equivalent to multiplying the two rotation matrices derived by the two quaternions. Since the quaternion $[1,0,0,0]$ corresponds to the identity rotation matrix, we have $\mathbf{q}_d=\mathbf{q}_c$ when $\delta\mathbf{q}=\mathbf{0}$.

\subsection{Rigid Transformation}
\label{sec:rigid}
We further transform the non-rigidly deformed 3D Gaussians $\{\mathcal{G}_d\}$ to the observation space via a rigid transformation module:
\begin{equation}
    \{\mathcal{G}_o\}=\mathcal{F}_{\theta_r}(\{\mathcal{G}_d\};\{\mathbf{B_b}\}_{b=1}^B)
    \label{eq:r_gaussians}
\end{equation}
where a skinning MLP $f_{\theta_r}$ is learned to predict skinning weights at the position $\mathbf{x}_d$. We transform the position and the rotation matrix of 3D Gaussians via forward LBS:
\begin{align}
    \mathbf{T}&=\sum\nolimits_{b=1}^B f_{\theta_r}(\mathbf{x}_d)_b\mathbf{B}_b \label{eq:r_gaussian_transform} \\
    \mathbf{x}_o&=\mathbf{T}\mathbf{x}_d \label{eq:r_gaussian_xyz}\\
    \mathbf{R}_o&=\mathbf{T}_{1:3,1:3}\mathbf{R}_d \label{eq:r_gaussian_rot}
\end{align}
where $\mathbf{R}_d$ is the rotation matrix derived from the quaternion $\mathbf{q}_d$. 

\subsection{Color MLP}
\label{sec:color}
Prior works~\cite{yang2023gs4d,wu20234dgaussians,yang2023deformable3dgs} follow the convention of 3DGS~\cite{kerbl3Dgaussians}, which stores spherical harmonics coefficients per 3D Gaussian to encode the view-dependent color. Treating the stored color feature $\mathbf{f}$ as spherical harmonics coefficients, the color of a 3D Gaussian can be computed by the dot product of the spherical harmonics basis and the learned coefficients:
$c=\langle\bm{\gamma}(\mathbf{d}),\mathbf{f}\rangle$,
%
%
where $\mathbf{d}$ represents the viewing direction, derived from the relative position of the 3D Gaussian wrt. the camera center and $\bm{\gamma}$ denotes the spherical harmonics basis function. 

While conceptually simple, we argue that this approach does not suit our monocular setting. Since only one camera view is provided during training, the viewing direction in the world space is fixed, leading to poor generalization to unseen test views. Similar to~\cite{peng2022animatable}, we use the inverse rigid transformation from \cref{sec:rigid} to canonicalize the viewing direction: 
$\hat{\mathbf{d}}=\mathbf{T}_{1:3,1:3}^{-1}\mathbf{d}$, 
%
%
where $\mathbf{T}$ is the forward transformation matrix defined in \cref{eq:r_gaussian_transform}. Theoretically, canonicalizing viewing direction also promotes consistency of the specular component of canonical 3D Gaussians under rigid transformations.

On the other hand, we observe that the pixel color of the rendered clothed human avatar also largely depends on local deformation. Local fine wrinkles on clothes, for instance, would cause self-occlusion that heavily affects shading. Following~\cite{peng2020neural}, we also learn a per-frame latent code $\mathcal{Z}_c$ to compensate for different environment light effects across frames caused by the global movement of the subject.
Hence, instead of learning spherical harmonic coefficients, we enhance color modeling by learning a neural network that takes per-Gaussian color feature vector $\mathbf{f}\in\mathbb{R}^{32}$, local pose-dependent feature vector $\mathbf{z}\in\mathbb{R}^{16}$ from the non-rigid deformation network, per-frame latent code $\mathcal{Z}_c\in\mathbb{R}^{16}$, and spherical harmonics basis of canonicalized viewing direction $\bm{\gamma}(\hat{\mathbf{d}})$ with a degree of $3$ as input and predicts the color of the 3D Gaussian:
\begin{equation}
    c=\mathcal{F}_{\theta_c}(\mathbf{f},\mathbf{z},\mathcal{Z}_c,\bm{\gamma}(\hat{\mathbf{d}}))
    \label{eq:color_mlp}
\end{equation}
In practice, we find a tiny MLP with one 64-dimension hidden layer sufficient to model the appearance. Increasing the size of the MLP leads to overfitting and performance drop.

\subsection{Optimization}
\label{sec:opt}
We jointly optimize canonical 3D Gaussians $\{\mathcal{G}_c\}$ and the parameters $\theta_{nr},\theta_r,\theta_c$ of the non-rigid deformation network, the skinning network and the color network, respectively.

\myparagraph{Pose correction.} 
SMPL~\cite{SMPL:2015} parameter fittings from images can be inaccurate. 
To address this, we additionally optimize the per-sequence shape parameter as well as per-frame translation, global rotation, and local joint rotations. We initialize these parameters $\theta_p$ with the given SMPL parameters and differentiably derive the bone transformations $\{\mathbf{B}_b\}$ as input to the network, enabling direct optimization via backpropagation.

\myparagraph{As-isometric-as-possible regularization.}
With monocular video as input, only one view of the human is visible in each frame, making it extremely hard to generalize to novel views and novel poses. Considering the sparsity of input, the non-rigid deformation network is highly underconstrained, resulting in noisy deformation from the canonical space to the observation space.
Inspired by ~\cite{Prokudin_2023_ICCV}, we leverage the as-isometric-as-possible constraint~\cite{kmp_shape_space_sig_07} to restrict neighboring 3D Gaussian centers to preserve a similar distance after deformation. We further augment the constraint to Gaussian covariance matrices:
\begin{align}
    \mathcal{L}_{isopos}=\sum_{i=1}^N \sum_{j\in\mathcal{N}_k(i)}\left|
    d(\mathbf{x}_c^{(i)},\mathbf{x}_c^{(j)})-d(\mathbf{x}_o^{(i)},\mathbf{x}_o^{(j)})
    \right|
    \label{eq:aiap_xyz}\\
    \mathcal{L}_{isocov}=\sum_{i=1}^N \sum_{j\in\mathcal{N}_k(i)}\left|
    d(\mathbf{\Sigma}_c^{(i)},\mathbf{\Sigma}_c^{(j)})-d(\mathbf{\Sigma}_o^{(i)},\mathbf{\Sigma}_o^{(j)})
    \right|
    \label{eq:aiap_cov}
\end{align}
where $N$ denotes the number of 3D Gaussians. $\mathcal{N}_k$ denotes the k-nearest neighbourhood, and we set $k$ to $5$. We use L2-norm as our distance function $d(\cdot,\cdot)$.

\myparagraph{Loss function.}
Our full loss function consists of a RGB loss $\mathcal{L}_{rgb}$, a mask loss $\mathcal{L}_{mask}$, a skinning weight regularization loss $\mathcal{L}_{skin}$ and the as-isometric-as-possible regularization loss for both position and covariance $\mathcal{L}_{isopos},\mathcal{L}_{isocov}$. For further details of the loss definition and respective weights, please refer to the Supp.Mat.

%% file: sec/5_experiments.tex
\section{Experiments}
\label{sec:exp}
In this section, we first compare the proposed approach with recent state-of-the-art methods~\cite{weng2022humannerf,ARAH:ECCV:2022,peng2020neural,instant_nvr,jiang2022instantavatar}, demonstrating that our proposed approach achieves superior rendering quality in terms of LPIPS, which is more informative under monocular setting, while achieving fast training and real-time rendering speed, respectively \textbf{400x and 250x faster} than the most competitive baseline~\cite{weng2022humannerf}. We then systematically ablate each component of the proposed model, showing their effectiveness in better rendering quality.

\subsection{Evaluation Dataset}
\label{sec:dataset}
\myparagraph{ZJU-MoCap~\cite{peng2020neural}.} This is the major testbed for quantitative evaluation. We pick six sequences (377, 386, 387, 392, 393, 394) from the ZJU-MoCap dataset and follow the training/test split of HumanNeRF~\cite{weng2022humannerf}. The motion of these sequences is repetitive and does not contain a sufficient number of poses for meaningful novel pose synthesis benchmarks. Thus we focus on evaluating novel view synthesis (PSNR/SSIM/LPIPS) and show qualitative results for animation on out-of-distribution poses. Note that LPIPS in all the tables are scaled up by $1000$.

\myparagraph{PeopleSnapshot~\cite{alldieck2018video}.} We also conduct experiments on 4 sequences of the PeopleSnapshot dataset, which includes monocular videos of people rotating in front of a camera. We follow the data split of InstantAvatar~\cite{jiang2022instantavatar} and compare to~\cite{jiang2022instantavatar} on novel pose synthesis. For fair comparison, we use the provided poses optimized by Anim-NeRF~\cite{peng2021animatable} and do not further optimize it during our training.

\begin{figure*}[t!]
    \begin{center}
        \includegraphics[width=0.92\linewidth]{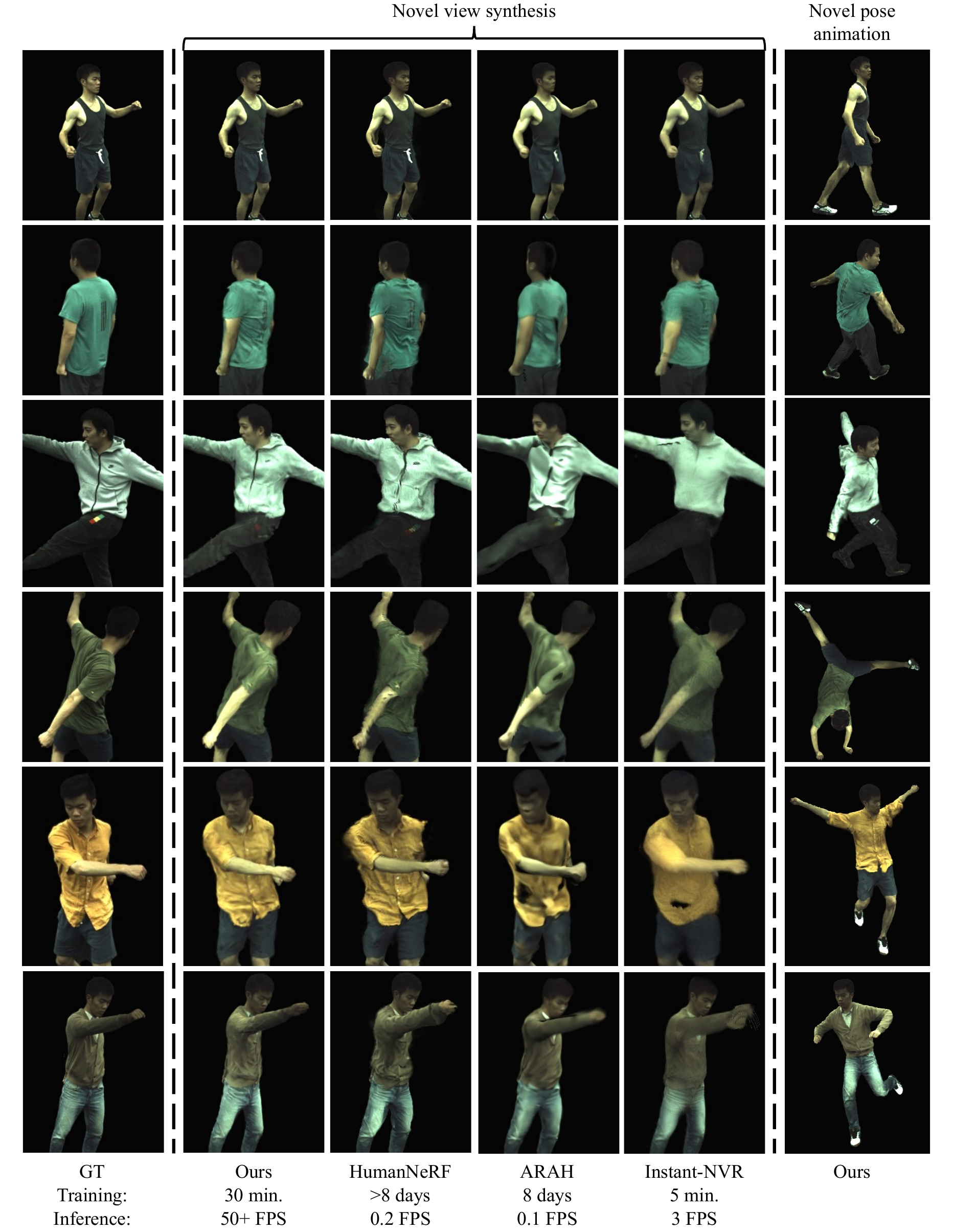}
    \end{center}
\caption{\textbf{Qualitative Comparison on ZJU-MoCap~\cite{peng2020neural}.}
We show the results for both novel view synthesis and novel pose animation of all sequences on ZJU-MoCap. Our method produces high-quality results that preserve cloth details even on out-of-distribution poses.
}
\label{fig:zju_qualitative}
\end{figure*}

\begin{table*}
\setlength{\fboxsep}{0pt}
\fontsize{6}{7}\selectfont
 \caption{\textbf{Quantitative Results on ZJU-MoCap~\cite{peng2020neural}.}
 We outperform both competitive baselines~\cite{weng2022humannerf,ARAH:ECCV:2022} in terms of LPIPS while being two orders of magnitude faster in training and rendering.  Cell color indicates \colorbox[RGB]{\colorfirst}{best} and \colorbox[RGB]{\colorsecond}{second best}.
 Instant-NVR~\cite{instant_nvr} is trained and tested on a refined version of ZJU-MoCap, thus is not directly comparable to other baselines quantitatively. We train our model on the refined dataset for fair quantitative comparison to Instant-NVR and the metrics are reported in the last two rows of the table.
 }
 \label{tab:compare_zjumocap}
 \centering
 \setlength{\tabcolsep}{2pt}
 \renewcommand{\arraystretch}{1.1}
 \begin{tabular}{ lcc|ccc|ccc|ccc|ccc|ccc|ccc}
 \toprule
 Subject:                                   
 & \multicolumn{2}{c}{}                                                  
 & \multicolumn{3}{c}{377}                                                 
 & \multicolumn{3}{c}{386}                                                 
 & \multicolumn{3}{c}{387}
 & \multicolumn{3}{c}{392}                   
 & \multicolumn{3}{c}{393}                    
 & \multicolumn{3}{c}{394}
 \\ 
Metric:                                     
& GPU$\downarrow$                    
& FPS$\uparrow$ 
& PSNR$\uparrow$
& SSIM$\uparrow$
& LPIPS$\downarrow$    
& PSNR$\uparrow$
& SSIM$\uparrow$
& LPIPS$\downarrow$    
& PSNR$\uparrow$
& SSIM$\uparrow$
& LPIPS$\downarrow$
& PSNR$\uparrow$
& SSIM$\uparrow$
& LPIPS$\downarrow$   
& PSNR$\uparrow$
& SSIM$\uparrow$
& LPIPS$\downarrow$    
& PSNR$\uparrow$
& SSIM$\uparrow$
& LPIPS$\downarrow$
 \\\hline
 NeuralBody~\cite{peng2020neural}           
 & \colorbox[RGB]{\colorsecond} {12h} 
 & \colorbox[RGB]{\colorsecond}{2} 
 & 29.11 
 & 0.9674 
 & 40.95                               
 & 30.54 
 & 0.9678 
 & 46.43                               
 & 27.00 
 & 0.9518 
 & 59.47 
 & 30.10 
 & 0.9642 
 & 53.27                                       
 & 28.61 
 & 0.9590 
 & 59.05                               
 & 29.10 
 & 0.9593 
 & 54.55 \\ 
 HumanNeRF~\cite{weng2022humannerf}         
 & $>$8d                             
 & 0.2                             
 & 30.41 
 & 0.9743 
 & \colorbox[RGB]{\colorsecond} {24.06}                               
 & 33.20 
 & 0.9752 
 & \colorbox[RGB]{\colorsecond}{28.99}                               
 & 28.18
 & 0.9632 
 & \colorbox[RGB]{\colorsecond}{35.58}  
 & 31.04 
 & 0.9705 
 & \colorbox[RGB]{\colorsecond}{32.12}                                       
 & 28.31 
 & 0.9603 
 & \colorbox[RGB]{\colorsecond}{36.72}                               
 & \colorbox[RGB]{\colorsecond}{30.31} 
 & \colorbox[RGB]{\colorsecond}{0.9642} 
 & \colorbox[RGB]{\colorsecond}{32.89} \\
 MonoHuman~\cite{yu2023monohuman}
 & 4d
 & 0.1
 & 29.12
 & 0.9727
 & 26.58
 & 32.94
 & 0.9695
 & 36.04
 & 27.93
 & 0.9601
 & 41.76
 & 29.50
 & 0.9635
 & 39.45
 & 27.64
 & 0.9566
 & 43.17
 & 29.15
 & 0.9595
 & 38.08
 \\
 ARAH~\cite{ARAH:ECCV:2022}                 
 & 8d
 & 0.1                             
 &\colorbox[RGB]{\colorfirst}{30.85}
 & \colorbox[RGB]{\colorfirst}{0.9800} 
 & 26.60  
 & \colorbox[RGB]{\colorsecond}{33.50} 
 & \colorbox[RGB]{\colorfirst}{0.9781} 
 & 31.40
 & \colorbox[RGB]{\colorfirst}{28.49} 
 & \colorbox[RGB]{\colorfirst}{0.9656} 
 & 40.43 
 & \colorbox[RGB]{\colorfirst}{32.02} 
 & \colorbox[RGB]{\colorfirst}{0.9742} 
 & 35.28
 & \colorbox[RGB]{\colorsecond}{28.77} 
 & \colorbox[RGB]{\colorfirst}{0.9645} 
 & 42.30
 & 29.46 
 & 0.9632 
 & 40.76 \\ 
 Ours                                      
 & \colorbox[RGB]{\colorfirst} {0.5h}  
 & \colorbox[RGB]{\colorfirst}{50} 
 & \colorbox[RGB]{\colorsecond}{30.64} 
 & \colorbox[RGB]{\colorsecond}{0.9774} 
 & \colorbox[RGB]{\colorfirst}{20.88} 
 & \colorbox[RGB]{\colorfirst}{33.63} 
 & \colorbox[RGB]{\colorsecond}{0.9773} 
 & \colorbox[RGB]{\colorfirst}{25.77} 
 & \colorbox[RGB]{\colorsecond}{28.33}
 & \colorbox[RGB]{\colorsecond}{0.9642} 
 & \colorbox[RGB]{\colorfirst}{34.24} 
 & \colorbox[RGB]{\colorsecond}{31.66} 
 & \colorbox[RGB]{\colorsecond}{0.9730} 
 & \colorbox[RGB]{\colorfirst}{30.14} 
 & \colorbox[RGB]{\colorfirst}{28.88} 
 & \colorbox[RGB]{\colorsecond}{0.9635} 
 & \colorbox[RGB]{\colorfirst}{35.26} 
 & \colorbox[RGB]{\colorfirst}{30.54} 
 & \colorbox[RGB]{\colorfirst}{0.9661} 
 & \colorbox[RGB]{\colorfirst}{31.21}
 \\ 
 \hline
 Instant-NVR*~\cite{instant_nvr}
 & \textbf{0.1h}
 & 3
 & \textbf{31.28} 
 & \textbf{0.9789} 
 & 25.37
 & 33.71 
 & 0.9770 
 & 32.81
 & 28.39 
 & 0.9640 
 & 45.97
 & 31.85 
 & 0.9730 
 & 39.47
 & \textbf{29.56} 
 & 0.9641 
 & 46.16
 & \textbf{31.32} 
 & \textbf{0.9680} 
 & 40.63 
 \\
  Ours*      
 & 0.5h
 & \textbf{50}
 & 30.96
 & 0.9778
 & \textbf{19.85}
 & \textbf{33.94}
 & \textbf{0.9784}
 & \textbf{24.70}
 & \textbf{28.40}
 & \textbf{0.9656}
 & \textbf{32.96}
 & \textbf{32.10}
 & \textbf{0.9739}
 & \textbf{29.20}
 & 29.30
 & \textbf{0.9645}
 & \textbf{34.03}
 & 30.74
 & 0.9662
 & \textbf{31.00}
 \\
 \bottomrule
 \end{tabular}
\end{table*}

\begin{table*}
\setlength{\fboxsep}{0pt}
\fontsize{9.5}{10.5}\selectfont
 \caption{\textbf{Quantitative Results on PeopleSnapshot~\cite{alldieck2018video}.}
 }
 \label{tab:compare_peoplesnapshot}
 \centering
 \setlength{\tabcolsep}{0.8pt}
 \renewcommand{\arraystretch}{1.1}
 \begin{tabular}{ lcc|ccc|ccc|ccc|ccc}
 \toprule
 Subject:                    
 & \multicolumn{2}{c}{}
 & \multicolumn{3}{c}{female-3-casual}                                                 
 & \multicolumn{3}{c}{female-4-casual}                                     
 & \multicolumn{3}{c}{male-3-casual}          
 & \multicolumn{3}{c}{male-4-casual}
 \\ 
Metric:             
& GPU$\downarrow$
& FPS$\uparrow$
& PSNR$\uparrow$
& SSIM$\uparrow$
& LPIPS$\downarrow$    
& PSNR$\uparrow$
& SSIM$\uparrow$
& LPIPS$\downarrow$    
& PSNR$\uparrow$
& SSIM$\uparrow$
& LPIPS$\downarrow$
& PSNR$\uparrow$
& SSIM$\uparrow$
& LPIPS$\downarrow$ 
 \\\hline
 InstantAvatar~\cite{jiang2022instantavatar}
 & \textbf{5 min.}
 & 15
 & 27.66
 & \textbf{0.9709}
 & 21.00    
 & 29.11
 & \textbf{0.9683}
 & 16.70     
 & 29.53
 & 0.9716
 & 15.50
 & 27.67
 & 0.9626
 & 30.7
 \\ 
 Ours  
 & 45 min.
 & \textbf{50}
 & \textbf{30.57}
 & 0.9581
 & \textbf{20.86}
 & \textbf{33.16}
 & 0.9678
 & \textbf{15.74}
 & \textbf{34.28}
 & \textbf{0.9724}
 & \textbf{14.92}
 & \textbf{30.22}
 & \textbf{0.9653}
 & \textbf{23.05}
 \\ 
 \bottomrule
  \end{tabular}
\end{table*}

\subsection{Comparison with Baselines}
\label{sec:comparison}
We compare our approach with NeuralBody~\cite{peng2020neural}, HumanNeRF~\cite{weng2022humannerf}, MonoHuman~\cite{yu2023monohuman}, ARAH~\cite{ARAH:ECCV:2022} and Instant-NVR~\cite{instant_nvr} under monocular setup on ZJU-MoCap. 
The quantitative results are reported in \cref{tab:compare_zjumocap}. 
NeuralBody is underperforming compared to other approaches. Overall, our proposed approach produces comparable performance to ARAH on PSNR and SSIM, while significantly outperforming all the baselines on LPIPS. We argue that LPIPS is more informative compared to the other two metrics, as it is very difficult to reproduce exactly the ground-truth appearance for novel views due to the monocular setting and the stochastic nature of cloth deformations. Meanwhile, our method is also capable of fast training and renders at a real-time rendering frame rate, being 400 times faster for training (30 GPU minutes \textit{vs.} 8 GPU days) and $250-500$ times faster for inference (50 FPS \textit{vs.} 0.1 FPS for ARAH and 0.2 FPS for HumanNeRF).
We also note that Instant-NVR trains on a refined version of ZJU-MoCap, which provides refined camera parameters, SMPL fittings, and more accurate instance masks with part-level annotation that is essential for running their method. Hence their metrics are not directly comparable to other methods in \cref{tab:compare_zjumocap}. We train our model on the refined dataset for a fair quantitative comparison, which clearly shows that our method outperforms Instant-NVR in most scenarios.

Qualitative comparisons on novel view synthesis can be found in \cref{fig:zju_qualitative}.
We observe that our method preserves sharper details compared to ARAH and does not produce fluctuating artifacts as in HumanNeRF caused by noisy deformation fields. 
Instant-NVR produces an oversmooth appearance and tends to generate noisy limbs.
Additionally, we animate our learned avatars with pose sequences from AMASS~\cite{AMASS:ICCV:2019} and AIST++~\cite{aist++:ICCV:2021}, shown in the rightmost column of \cref{fig:zju_qualitative}. This shows that our model could generalize to extreme out-of-distribution poses. 

For PeopleSnapshot, we report the quantitative comparison against InstantAvatar~\cite{jiang2022instantavatar} in \cref{tab:compare_peoplesnapshot}. Our approach significantly outperforms InstantAvatar on PSNR and LPIPS, while being more than 3x faster during inference.

\begin{table}
 \caption{\textbf{Ablation Study on ZJU-MoCap~\cite{peng2020neural}.} 
 The proposed model achieves the lowest LPIPS, demonstrating the effectiveness of all components.}
 \label{tab:ablation}
 \centering
 \begin{tabular}{@{}lccc}
 \toprule
 Metric:          
 & PSNR$\uparrow$
 & SSIM$\uparrow$
 & LPIPS$\downarrow$\\ \hline
 Full model               
 & \textbf{30.61}
 & \textbf{0.9703}
 & \textbf{29.58}
\\
  w/o color MLP
 & 30.55
 & 0.9700
 & 31.24
\\
  w/o $\mathcal{L}_{isocov}$ 
 &  \textbf{30.61}
 & \textbf{0.9703}
 & 29.84
  \\
  w/o $\mathcal{L}_{isopos}, \mathcal{L}_{isocov}$      
 & 30.59
 & 0.9699
 & 30.25
 \\
 w/o pose correction  
 & 30.60
 & \textbf{0.9703}
 & 29.87
  \\
 \bottomrule
 \end{tabular}
\end{table}

\begin{figure*}[h!]
    \begin{center}
        \includegraphics[width=0.95\linewidth]{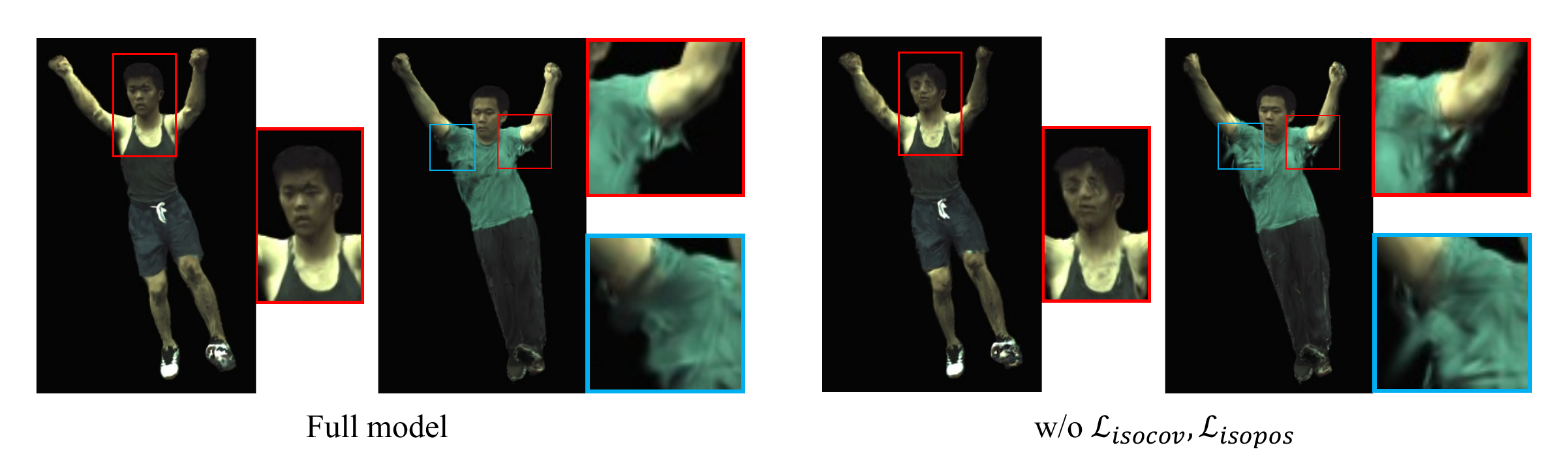}
    \end{center}
\caption{\textbf{Ablation Study} on as-isometric-as-possible regularization, which removes the artifacts on highly articulated poses.
}
\label{fig:aiap_ablation}
\end{figure*}

\subsection{Ablation Study}
\label{sec:ablation}

We study the effect of various components of our method on the ZJU-MoCap dataset, including the color MLP, the as-isometric-as-possible regularization and the pose correction module. The average metrics over $6$ sequences are reported in \cref{tab:ablation}. We show that all proposed techniques are required to reach the optimal performance, best reflected by LPIPS which is the most informative metric for novel view synthesis evaluation under a monocular setup.

We further show qualitative comparison on out-of-distribution poses in \cref{fig:aiap_ablation}, which demonstrates that the as-isometric-as-possible loss helps to constrain the 3D Gaussians to comply with consistent movement during deformation, hence improving generalization on novel poses. Albeit marginally, each individual component contributes to a better novel-view rendering quality and particularly generates more plausible results with respect to novel pose animation. 

%% file: sec/6_conclusion.tex
\section{Conclusion}
\label{sec:conclusion}

In this paper, we present 3DGS-Avatar, one of the first methods that utilize the explicit representation of 3DGS for efficient reconstruction of clothed human avatars from monocular videos. Our method achieves photorealistic rendering, awareness of pose-dependent cloth deformation, generalization to unseen poses, fast training, and real-time rendering all at once.

Experiments show that our method is comparable to or even better than the state-of-the-art methods in terms of rendering quality while being two orders of magnitude faster in both training and inference. Furthermore, we propose to replace spherical harmonics with a shallow MLP to decode 3D Gaussian color and regularize deformation with geometric constraints, both proved to be effective in enhancing rendering quality. We hope that our new representation could foster further research in fast, high-quality animatable clothed human avatar synthesis from a monocular view.

\myparagraph{Acknowledgement.} SW and AG were supported by the ERC Starting Grant LEGO-3D (850533) and the DFG EXC number 2064/1 - project number 390727645. SW and ST acknowledge the SNSF grant 200021 204840. 

%% file: sec/X_suppl.tex
\clearpage
\setcounter{page}{1}
\maketitlesupplementary
\appendix

\input{sec/supplement/loss}
\input{sec/supplement/implementation}
\input{sec/supplement/baselines}
\input{sec/supplement/ablation}
\input{sec/supplement/qualitative}
\input{sec/supplement/limitation}

%% file: sec/supplement/loss.tex
\section{Loss Definition}
\label{sec:supp_loss}

In 
\cref{sec:opt} 
of the main paper we describe our loss term which can be formulated as follows:
\begin{equation}
    \mathcal{L}=
    \lambda_{l1}\mathcal{L}_{l1}+
    \lambda_{perc}\mathcal{L}_{perc}+
    \lambda_{mask}\mathcal{L}_{mask}+
    \lambda_{skin}\mathcal{L}_{skin}+
    \lambda_{isopos}\mathcal{L}_{isopos}+
    \lambda_{isocov}\mathcal{L}_{isocov}
\label{eq:loss}
\end{equation}
We describe how each loss term is defined below:

\paragraph{RGB Loss:}
We use an $l1$ loss to compute pixel-wise error and a perceptual loss to provide robustness to local misalignments, which is critical for the monocular setup. Following~\cite{weng2022humannerf}, we optimize LPIPS as the perceptual loss with VGG as the backbone. However, unlike NeRF-based methods which train on random ray samples, we render the whole image via rasterization and thus do not require patch sampling. For computational efficiency, we crop the tight enclosing bounding box with the ground truth mask and compute the VGG-based LPIPS as our perceptual loss.

\paragraph{Mask Loss:}
To boost the convergence of 3D Gaussian positions, we use an explicit mask loss. For each pixel $p$, we compute the opacity value $O_p$ by summing up the sample weights in the rendering equation 
\cref{eq:render} 
in the main paper
, namely:
\begin{equation}
    O_p=\sum\nolimits_i\alpha'_i\prod\nolimits_{j=1}^{i-1}(1-\alpha'_j)
    \label{eq:render_opacity}
\end{equation}
We thus supervise it with the ground truth foreground mask via an $l1$ loss. Experiments show that the $l1$ loss provides faster convergence than the Binary Cross Entropy (BCE) loss.

\paragraph{Skinning Loss:} 
We leverage SMPL prior by sampling $1024$ points $\mathbf{X}_{skin}$ on the surface of the canonical SMPL mesh and regularizing the forward skinning network with corresponding skinning weights $\mathbf{w}$ interpolated with barycentric coordinates.
\begin{equation}
    \mathcal{L}_{skin}=\frac{1}{|\mathbf{X}_{skin}|}\sum_{\mathbf{x}_{skin}\in\mathbf{X}_{skin}}||f_{\theta_r}(\mathbf{x}_{skin})-\mathbf{w}||^2
\end{equation}

\paragraph{As-isometric-as-possible Loss:} Please refer to the second paragraph of 
\cref{sec:opt} 
in the main paper
for details.

We set $\lambda_{l1}=1,\lambda_{perc}=0.01, \lambda_{mask}=0.1,\lambda_{isopos}=1,\lambda_{isocov}=100$ in all experiments. For $\lambda_{skin}$, we set it to $10$ for the first $1k$ iterations for fast convergence to a reasonable skinning field, then decreased to $0.1$ for soft regularization.

%% file: sec/supplement/implementation.tex
\section{Implementation Details}
\label{sec:supp_implementation}

We initialize the canonical 3D Gaussians with $N=50k$ random samples on the SMPL mesh surface in canonical pose. During optimization, we follow the same strategy from~\cite{kerbl3Dgaussians} to densify and prune the 3D Gaussians, using the view-space position gradients derived from the transformed Gaussians $\mathcal{G}_o$ in the observation space as the criterion for densification.

We then describe the network architectures of our learned neural components.
For the forward skinning network $f_{\theta_r}$, we use an MLP with $4$ hidden layers of $128$ dimensions which takes $\mathbf{x}_c\in\mathbb{R}^3$ with no positional encoding and outputs a $25$-dimension vector. This vector is further propagated through a hierarchical softmax layer that is aware of the tree structure of the human skeleton to obtain the skinning weights $\mathbf{w}$ that sum up to $1$. To normalize the coordinates in the canonical space, we proportionally pad the bounding box enclosing the canonical SMPL mesh instead of using the same length in all axes as in~\cite{ARAH:ECCV:2022}. This allows us to use a lower resolution in the flat $z$-dimension of the human body.

For the non-rigid deformation network $f_{\theta_{nr}}$, the 3D position $\mathbf{x}_d$ is normalized with the aforementioned bounding box and first encoded into representative features with a multi-level hash grid, whose parameters are defined in \cref{tab:ingp}. The concatenation of the hash grid features and the pose latent code $\mathcal{Z}_p$ then go through a shallow MLP with $3$ hidden layers of $128$ dimensions to decode pose-dependent local deformation.

The details of our color network structure $\mathcal{F}_{\theta_c}$ are well elaborated in 
\cref{sec:color} 
of the main paper.
For frames outside the training set, we follow~\cite{ARAH:ECCV:2022} and use the latent code of the last frame in the training sequence.

\begin{table}[]
    \centering
    \begin{tabular}{c|c}
    \hline
     Parameter & Value \\
     \hline
      Number of levels & 16 \\
      Feature dimension per level  & 2\\
      Hash table size & $2^{16}$ \\
      Coarsest resolution & 16 \\
      Finest resolution & 2048 \\
      \hline
    \end{tabular}
    \caption{Hash table parameters.}
    \label{tab:ingp}
\end{table}

To reduce overfitting, we add noise to the pose and viewing direction input. Specifically, we add a noise drawn from the normal distribution $\mathcal{N}(0,0.1)$ to the SMPL pose parameters $\theta$ with a probability of $p=0.5$ during training. The viewing direction $d$ is first canonicalized to the canonical space and then augmented with a random rotation derived from uniformly sampled roll, pitch, and yaw degrees $\in[0,45)$.  Adding noise to training signals helps the model to better generalize to novel poses and views.

Our model is trained for a total of $15k$ iterations on the ZJU-MoCap dataset in $30$ minutes and $30k$ iterations on PeopleSnapshot in $45$ minutes on a single NVIDIA RTX 3090 GPU. We use Adam~\cite{Adam:ICLR:2015} to optimize our model and the per-frame latent codes with hyperparameters $\beta_1=0.9$ and $\beta_2=0.999$. The learning rate of 3D Gaussians is exactly the same as the original implementation from~\cite{kerbl3Dgaussians}. We set the learning rate for forward skinning network $\theta_r$ to $1\times 10^{-4}$ and $1\times 10^{-3}$ for all the others. An exponential learning rate scheduler is employed to gradually decrease the learning rate by a factor of $0.1$ on neural networks. We also apply a weight decay with a weight of $0.05$ to the per-frame latent codes.

Following prior works~\cite{weng2022humannerf, yang2023deformable3dgs}, we split the training stage and learn the whole model in a coarse-to-fine manner. In the first $1k$ iterations, we freeze everything except the forward skinning network $f_{\theta_r}$ to learn a coarse skinning field with $\mathcal{L}_{skin}$ and prevent the noisy gradients from moving the 3D Gaussians away from the initialization. We then enable optimization on the 3D Gaussians after $1k$ steps. To decouple rigid and non-rigid motion, we start to optimize the non-rigid deformation network $f_{\theta_{nr}}$ after $3k$ iterations. Lastly, we turn on pose correction after $5k$ iterations.

%% file: sec/supplement/baselines.tex
\section{Implementation Details for Baselines}
\label{sec:supp_baseline}

In this section, we elaborate on the implementation details of baselines used for comparison to our proposed method, \textit{i.e.} NeuralBody~\cite{peng2020neural}, HumanNeRF~\cite{weng2022humannerf}, ARAH~\cite{ARAH:ECCV:2022}, Instant-NVR~\cite{instant_nvr}, MonoHuman~\cite{yu2023monohuman} and InstantAvatar~\cite{jiang2022instantavatar}.

\subsection{NeuralBody}
For the quantitative evaluation, we use the results of NeuralBody~\cite{peng2020neural}  reported in  HumanNeRF~\cite{weng2022humannerf} which follows the same data split.

\subsection{HumanNeRF}
We use pre-trained models provided by the official code repository\footnote{https://github.com/chungyiweng/humannerf} for both quantitative and qualitative evaluation. 

\subsection{ARAH}
For the quantitative evaluation, we use the same setup as HumanNeRF (\ie\ same data split with a reduced image size of $512 \times 512$) and train the models using the code from official code repository\footnote{https://github.com/taconite/arah-release} for 500 epochs. All other hyperparameters remain unchanged. The trained models are then used for qualitative evaluation and out-of-distribution pose animation.

\subsection{Instant-NVR}
For quantitative and qualitative evaluation, we retrain the models using the code from official code repository\footnote{https://github.com/zju3dv/instant-nvr} on the refined ZJU-MoCap dataset provided by the author. We change the data split to match other baselines while keeping all other hyperparameters the same. 

\subsection{MonoHuman}
We note that MonoHuman uses a different data split from HumanNeRF with the last fifth of the training frames being used for novel pose synthesis evaluation instead. For fair comparison we retrain the model from official code repository\footnote{https://github.com/Yzmblog/MonoHuman} on the same data split of HumanNeRF with the provided configs for $400k$ iterations and recompute the metrics on novel view synthesis. The trained models are then used for qualitative evaluation and out-of-distribution pose animation.

\subsection{InstantAvatar}
We follow the original setup and use the provided poses optimized by Anim-NeRF~\cite{peng2021animatable} without further pose correction. For quantitative results we copy the metrics from their table, while for qualitative results we train the model from official code repository\footnote{https://github.com/tijiang13/InstantAvatar} as they do not release pretrained checkpoints.

%% file: sec/supplement/ablation.tex
\section{Ablation Study}
\label{sec:supp_ablation}

We conduct additional ablation study and report the average metrics on ZJU-MoCap in \cref{tab:supp_ablation}.

\begin{table}
 \caption{\textbf{Additional Ablation Study on ZJU-MoCap~\cite{peng2020neural}.} 
 We present the average metrics over 6 sequences.
 }
 \label{tab:supp_ablation}
 \centering
 \begin{tabular}{@{}lccc}
 \toprule
 Metric:          
 & PSNR$\uparrow$
 & SSIM$\uparrow$
 & LPIPS$\downarrow$\\ \hline
 Full model               
 & \textbf{30.61}
 & \textbf{0.9703}
 & \textbf{29.58}
\\
 w/o $\mathcal{L}_{mask}$  
 & 30.58
 & \textbf{0.9703}
 & 29.90
\\
 Random initialization
 & \textbf{30.61}
 & 0.9701
 & 30.90
  \\
  $7k$ iterations
 & 30.56
 & 0.9698
 & 31.73
  \\
 \bottomrule
 \end{tabular}
\end{table}

\subsection{Ablation on Network Components}
To showcase the effect of each MLP component in our model on both training efficiency and quality, we additionally ablate  respective network-free variants: (1) shallow color MLP $\mathcal{F}_{\theta_c}$ is replaced by spherical harmonics function, (2) no non-rigid deformation $\mathcal{F}_{\theta_{nr}}$, (3) learned skinning field $\mathcal{F}_{\theta_r}$ is replaced by querying the skinning weight of the nearest SMPL vertex. The results are shown in Tab.~\ref{tab:supp_ablation_network}. We surprisingly find that using the SMPL nearest neighbor skinning does not harm the quality while further reducing the training time on ZJU-MoCap dataset. The result is not sensitive to the skinning field possibly due to subsequent compensation of non-rigid deformation. However, we keep to learn the skinning field for its flexibility and generalization to diverse clothing.

\begin{table}[ht!]
    \fontsize{10}{11}\selectfont
     \centering
     \setlength{\tabcolsep}{5pt}
     \begin{tabular}{@{}l|c|c|c|c|c}
     \toprule    
     & Full
     & (1)
     & (2)
     & (3)
     & (1)(2)(3)
     \\ \hline
     \addlinespace[2pt]
     Time   
     & 0:24
     & 0:24
     & 0:20
     & 0:19
     & 0:12
    \\
    LPIPS         
     & 29.58
     & 31.24
     & 32.31
     & 29.54
     & 32.67
    \\
     \bottomrule
     \end{tabular}
     \caption{\textbf{Balance between quality and efficiency.} 
     We present the average LPIPS over 6 sequences and the respective training time under each setting. 
     }
     \label{tab:supp_ablation_network}
\end{table}

\subsection{Ablation on Color MLP}
\begin{figure*}[ht!]
    \begin{center}
        \includegraphics[width=0.8\linewidth]{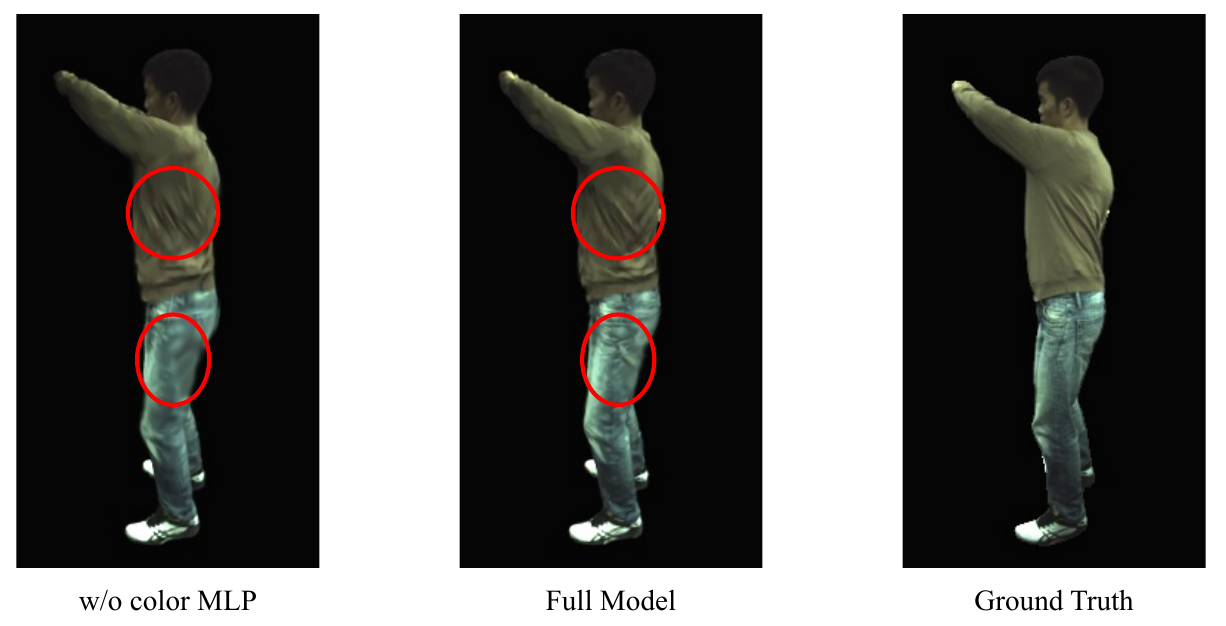}
    \end{center}
\caption{\textbf{Qualitative Ablation of Color MLP.} 
}
\label{fig:supp_ablation_tex}
\end{figure*}

We show in 
\cref{tab:ablation}
of the main paper that our proposed color MLP produces rendering with higher quality compared to learning spherical harmonics coefficients. We hereby show qualitative comparison to corroborate this enhancement in \cref{fig:supp_ablation_tex}. Our proposed color MLP helps generate more realistic cloth wrinkles and sharper textures with pose-dependent feature $\mathbf{z}$ and per-frame latent code $\mathcal{Z}_c$ as additional inputs.

\subsection{Ablation on Pose Correction}
\begin{figure}[ht!]
  \centering
  \includegraphics[width=0.6\linewidth]{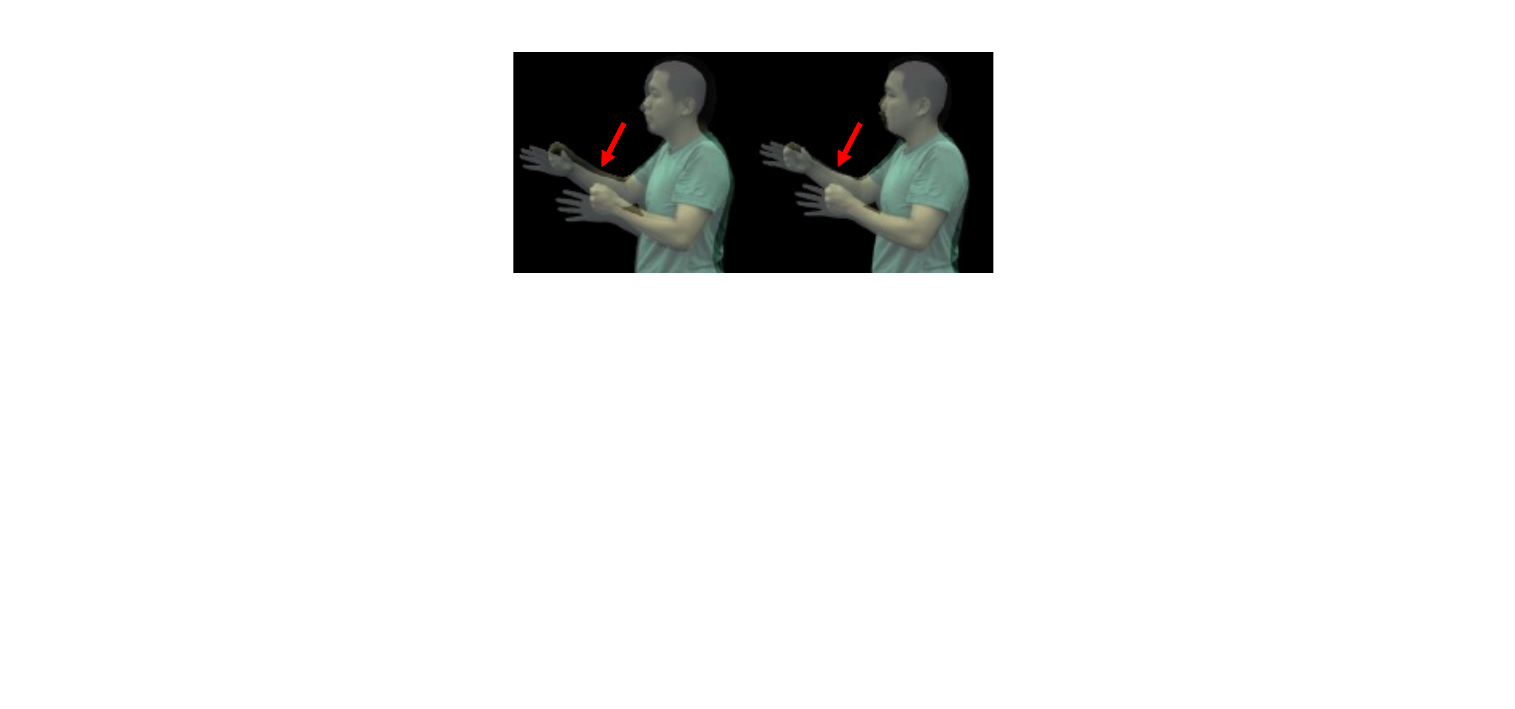}
   \caption{
   \textbf{Qualitative Ablation of Pose Correction.}
   \textit{left:} before pose correction, \textit{right:} after pose correction. 
   The SMPL mesh aligns better with the ground-truth image after pose optimization.
   }
   \label{fig:supp_ablation_pc}
\end{figure}

We additionally show the visualization of pose correction in Fig.~\ref{fig:supp_ablation_pc}. Following ARAH and HumanNeRF, we refine the inaccurate SMPL estimation during training, which helps improve the quality of avatar modeling.

\subsection{Additional Ablation on AIAP Regularization}
\begin{figure}[ht!]
    \centering
    \includegraphics[width=0.8\linewidth]{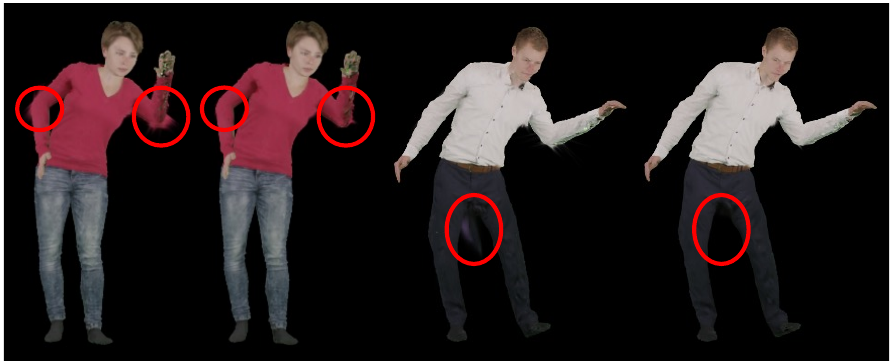}
     \caption{
     \textbf{Qualitative Ablation of AIAP regularization on PeopleSnapshot.}
     For each subject, \textit{left:} w/o AIAP loss, \textit{right:} w/ AIAP loss. Red circles highlight where Gaussian deformations become noisy without enforcing the AIAP constraint.
     }
     \label{fig:supp_ablation_aiap}
\end{figure}

While forward LBS naturally generalizes to out-of-distribution poses, the pose-dependent non-rigid deformation module can be underconstrained and noisy without proper regularization.
To improve generalization, AIAP loss enforces local consistent deformation of Gaussians, thus removing scattered Gaussian artifacts away from the human body.
Similar effects can also be observed in novel pose synthesis results on PeopleSnapshot, as shown in Fig.~\ref{fig:supp_ablation_aiap}. While the AIAP loss shows marginal improvement on novel-view synthesis benchmark, it helps stabilize the Gaussian position and shape on unseen poses. 

\subsection{Ablation on Mask Supervision}
\begin{figure*}[ht!]
    \begin{center}
        \includegraphics[width=0.8\linewidth]{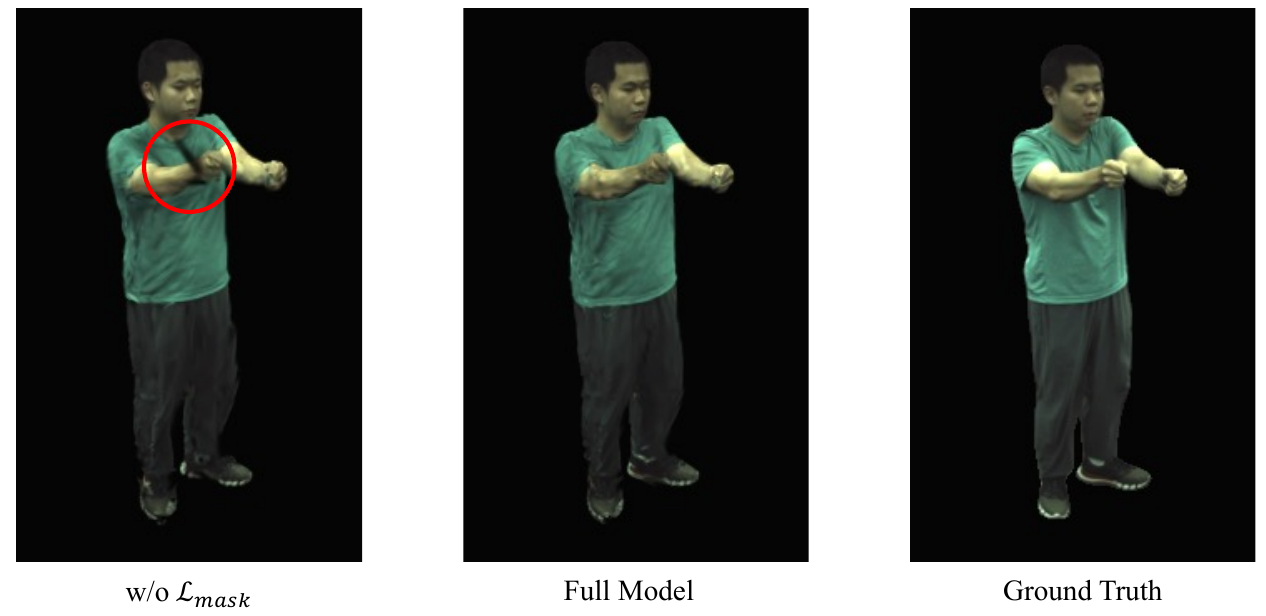}
    \end{center}
\caption{\textbf{Qualitative Ablation of Mask Loss.}
}
\label{fig:supp_ablation_mask}
\end{figure*}

Explicit supervision from ground-truth foreground masks only seems to gain slight improvement, as shown in \cref{tab:supp_ablation}. However, we observe that the mask loss is useful for removing floating blobs in the empty space. \cref{fig:supp_ablation_mask} shows an example for this, without mask loss, the floating Gaussian with the background color could occlude the subject in novel views.

\subsection{Ablation on Gaussian Initialization}

Instead of initializing the canonical 3D Gaussians from a SMPL mesh surface, we tried to perform random initialization. Specifically, we randomly sample $N=50k$ points in the enclosing bounding box around the canonical SMPL mesh. Experimental results from \cref{tab:supp_ablation} demonstrate that our method could as well converge starting from random initialization, with little performance drop compared to the SMPL initialization scheme. Despite this interesting observation, we decide to use SMPL initialization as it is more intuitive and does not incur any overhead.

\subsection{Ablation on Training Iterations}
\begin{figure*}[ht!]
    \begin{center}
        \includegraphics[width=0.8\linewidth]{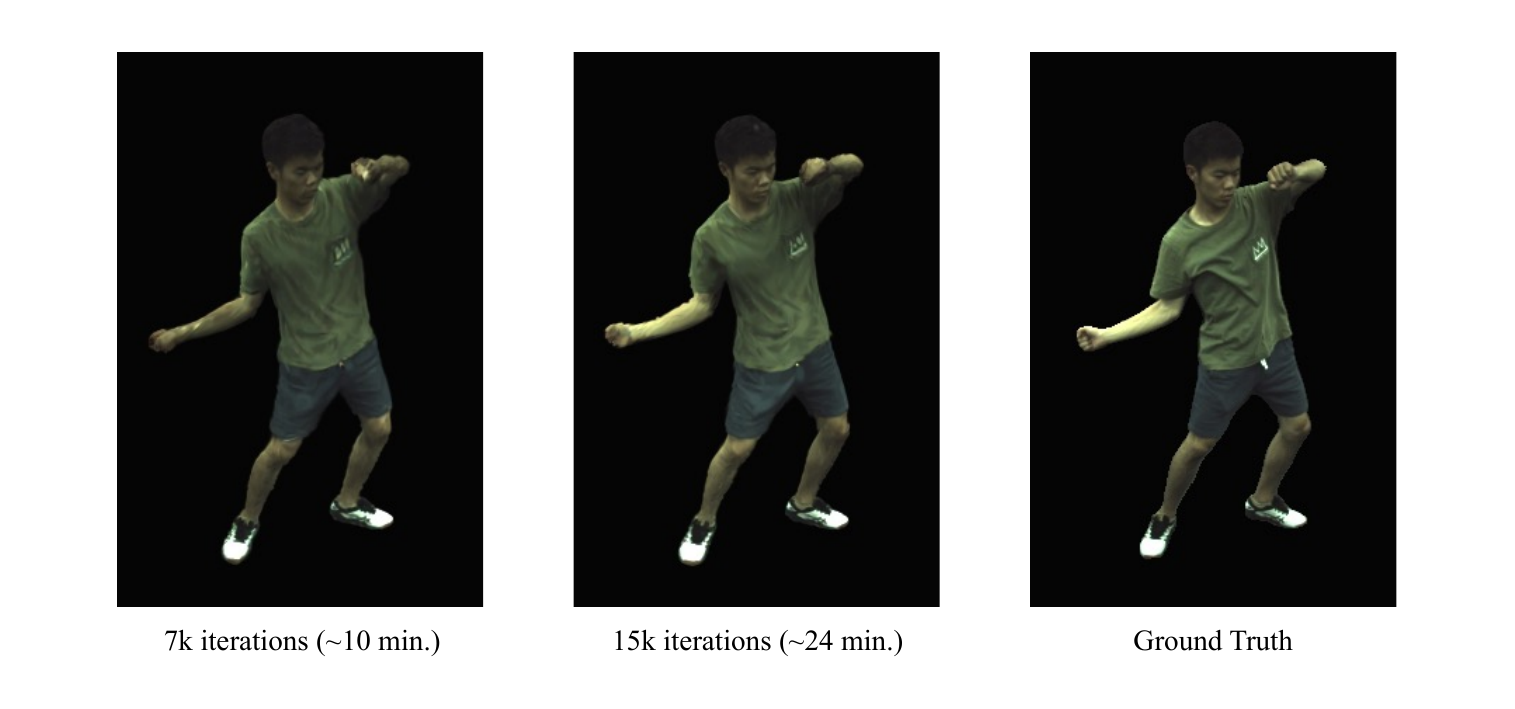}
    \end{center}
\caption{\textbf{Qualitative Ablation of Training Iterations.}
}
\label{fig:supp_ablation_7k}
\end{figure*}

Training for $15k$ iterations on ZJU-MoCap takes precisely around $24$ minutes. We further show that our method can already achieve high-quality results at $7k$ iterations in \cref{tab:supp_ablation}, which takes around $10$ minutes, not far away from \cite{jiang2022instantavatar} and \cite{instant_nvr} that claim instant training within $5$ minutes. Qualitative comparison is shown in \cref{fig:supp_ablation_7k}.

%% file: sec/supplement/qualitative.tex
\section{Additional Qualitative Results}
\label{sec:supp_qualitative}

We show more qualitative results in this section. \textbf{For better visualization, we strongly recommend to check our supplementary video.}

\begin{figure}[t!]
\captionsetup[subfigure]{labelformat=empty}
\scriptsize
\centering
\begin{subfigure}[b]{0.16\textwidth}
    \includegraphics [trim=150 180 150 80, clip, width=1.0\textwidth]{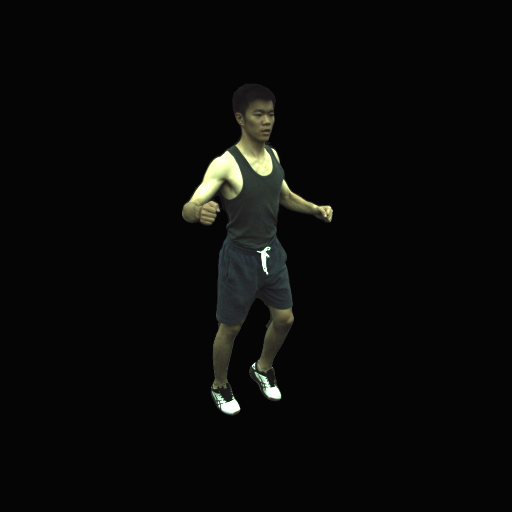}
\end{subfigure}
\begin{subfigure}[b]{0.16\textwidth}
    \includegraphics [trim=150 180 150 80, clip, width=1.0\textwidth]{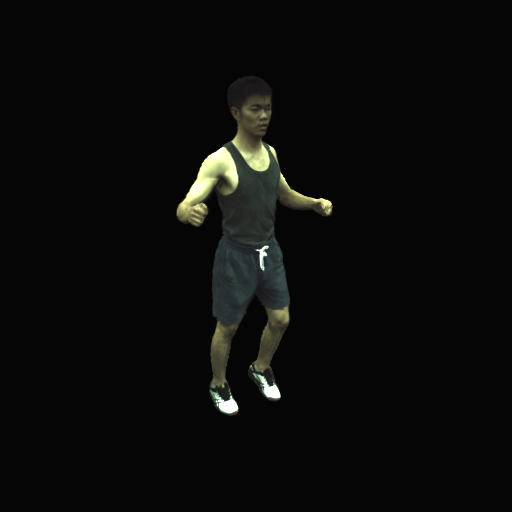}
\end{subfigure}
\begin{subfigure}[b]{0.16\textwidth}
    \includegraphics [trim=150 180 150 80, clip, width=1.0\textwidth]{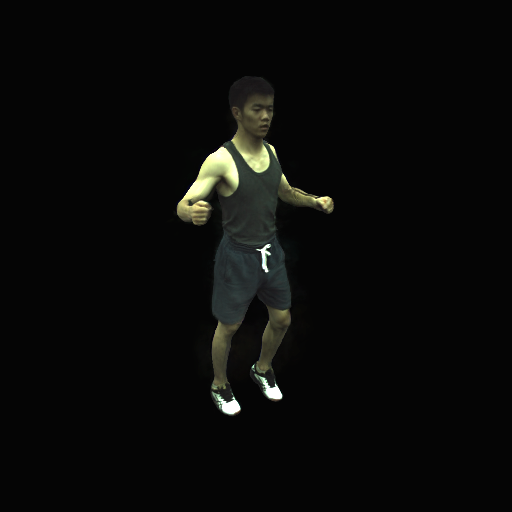}
\end{subfigure}
\begin{subfigure}[b]{0.16\textwidth}
    \includegraphics [trim=150 180 150 80, clip, width=1.0\textwidth]{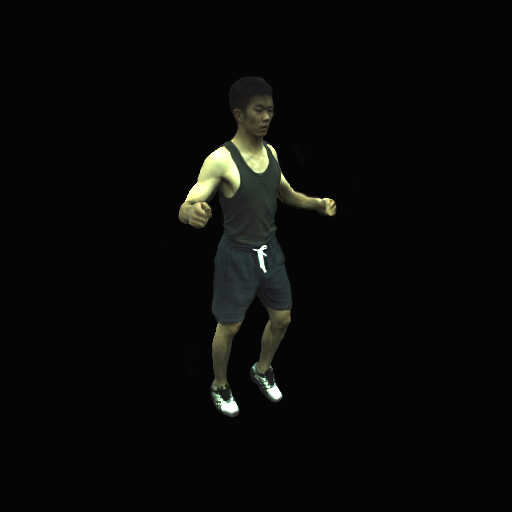}
\end{subfigure}
\begin{subfigure}[b]{0.16\textwidth}
    \includegraphics [trim=150 180 150 80, clip, width=1.0\textwidth]{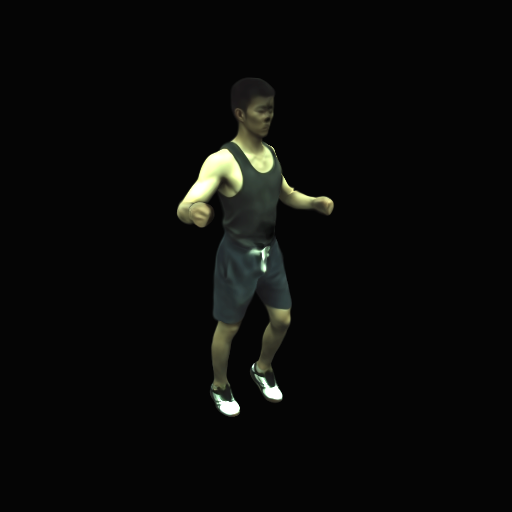}
\end{subfigure}
\begin{subfigure}[b]{0.16\textwidth}
    \includegraphics [trim=150 180 150 80, clip, width=1.0\textwidth]{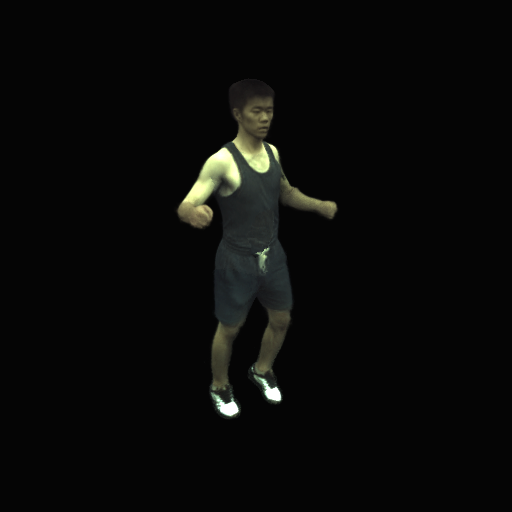}
\end{subfigure}
\\
\begin{subfigure}[b]{0.16\textwidth}
    \includegraphics [trim=30 180 280 80, clip, width=1.0\textwidth]{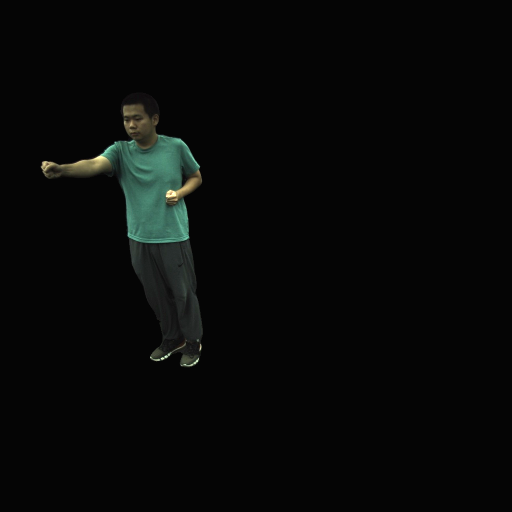}
\end{subfigure}
\begin{subfigure}[b]{0.16\textwidth}
    \includegraphics [trim=30 180 280 80, clip, width=1.0\textwidth]{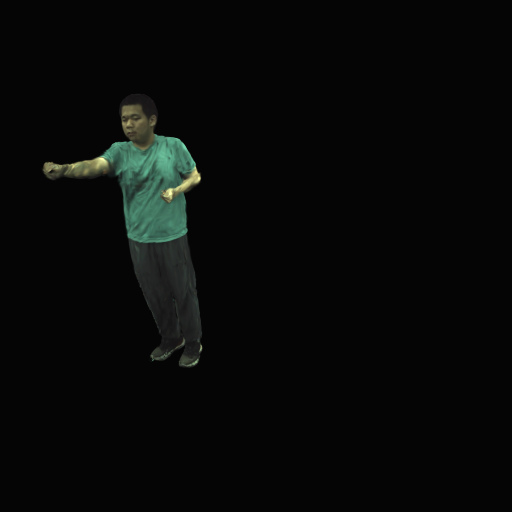}
\end{subfigure}
\begin{subfigure}[b]{0.16\textwidth}
    \includegraphics [trim=30 180 280 80, clip, width=1.0\textwidth]{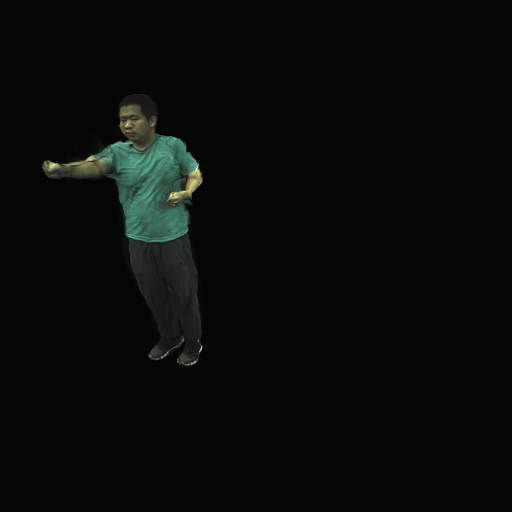}
\end{subfigure}
\begin{subfigure}[b]{0.16\textwidth}
    \includegraphics [trim=30 180 280 80, clip, width=1.0\textwidth]{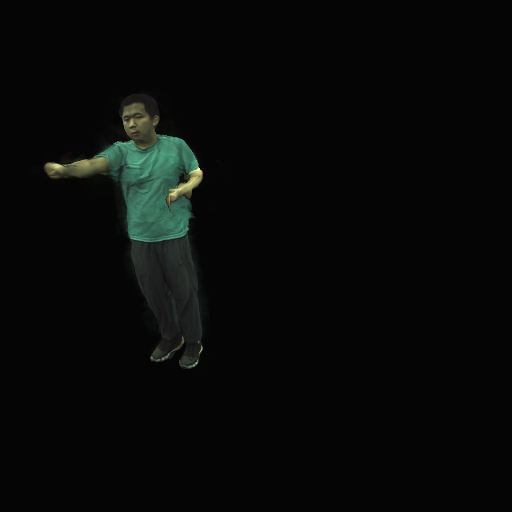}
\end{subfigure}
\begin{subfigure}[b]{0.16\textwidth}
    \includegraphics [trim=30 180 280 80, clip, width=1.0\textwidth]{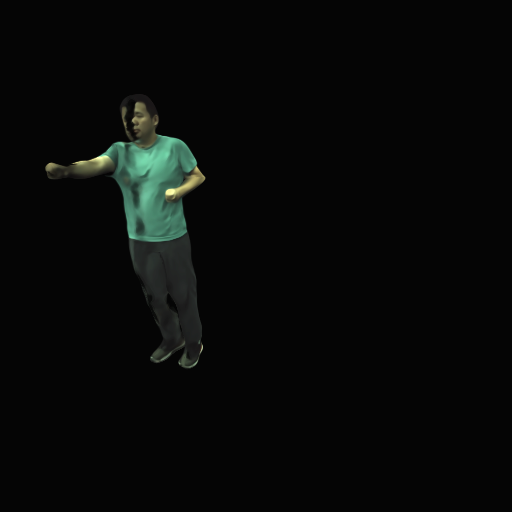}
\end{subfigure}
\begin{subfigure}[b]{0.16\textwidth}
    \includegraphics [trim=30 180 280 80, clip, width=1.0\textwidth]{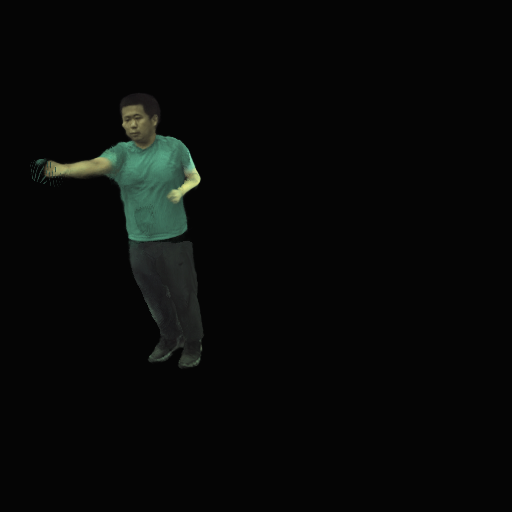}
\end{subfigure}
\\
\begin{subfigure}[b]{0.16\textwidth}
    \includegraphics [trim=150 180 200 80, clip, width=1.0\textwidth]{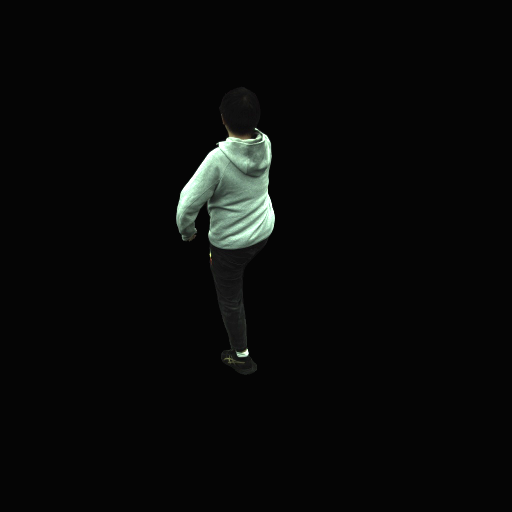}
\end{subfigure}
\begin{subfigure}[b]{0.16\textwidth}
    \includegraphics [trim=150 180 200 80, clip, width=1.0\textwidth]{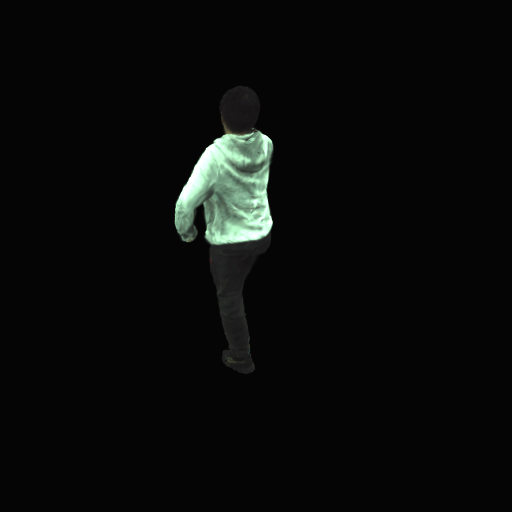}
\end{subfigure}
\begin{subfigure}[b]{0.16\textwidth}
    \includegraphics [trim=150 180 200 80, clip, width=1.0\textwidth]{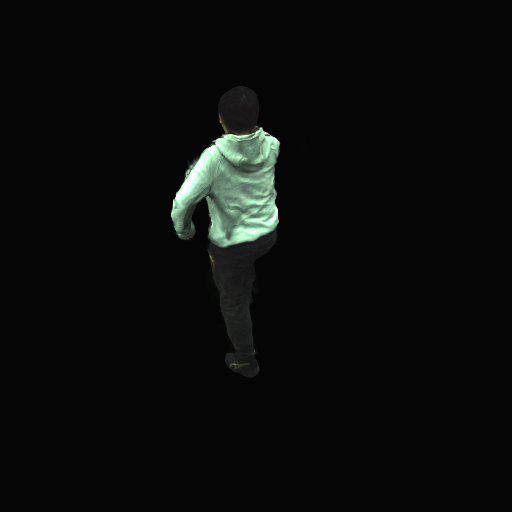}
\end{subfigure}
\begin{subfigure}[b]{0.16\textwidth}
    \includegraphics [trim=150 180 200 80, clip, width=1.0\textwidth]{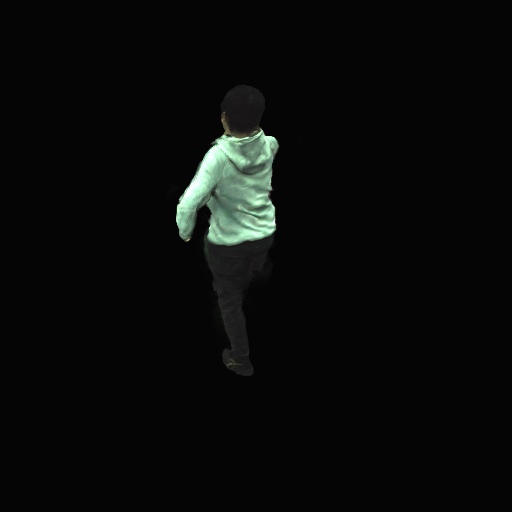}
\end{subfigure}
\begin{subfigure}[b]{0.16\textwidth}
    \includegraphics [trim=150 180 200 80, clip, width=1.0\textwidth]{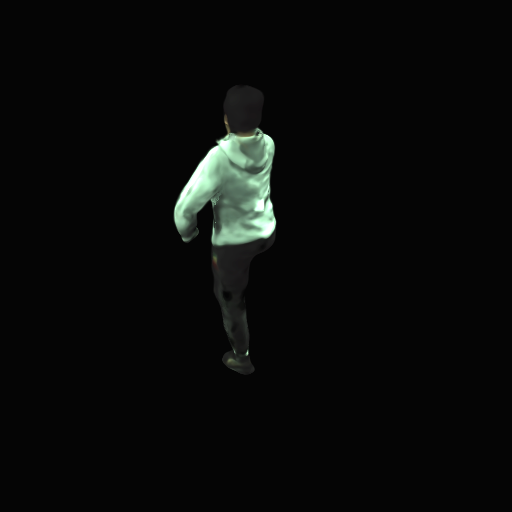}
\end{subfigure}
\begin{subfigure}[b]{0.16\textwidth}
    \includegraphics [trim=150 180 200 80, clip, width=1.0\textwidth]{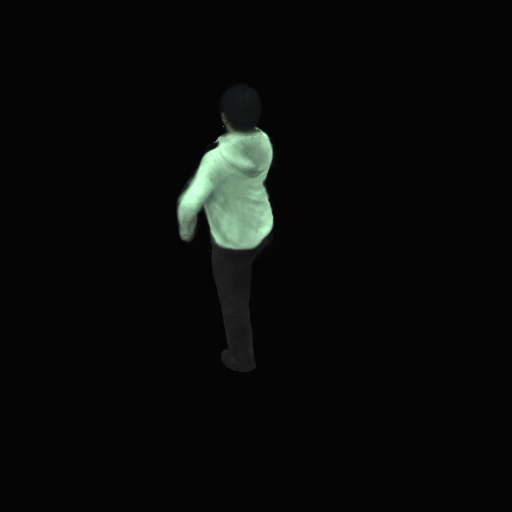}
\end{subfigure}
\\
\begin{subfigure}[b]{0.16\textwidth}
    \includegraphics [trim=150 180 150 80, clip, width=1.0\textwidth]{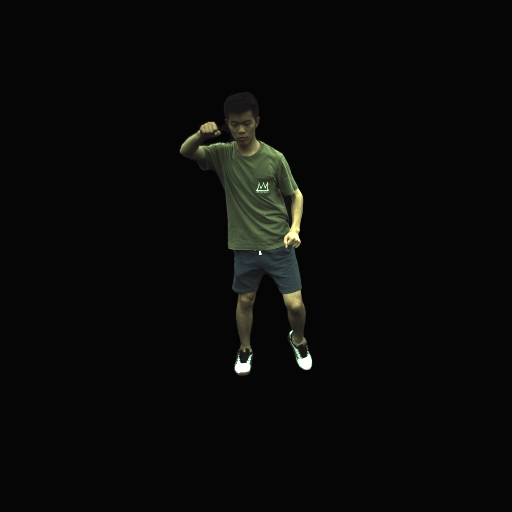}
\end{subfigure}
\begin{subfigure}[b]{0.16\textwidth}
    \includegraphics [trim=150 180 150 80, clip, width=1.0\textwidth]{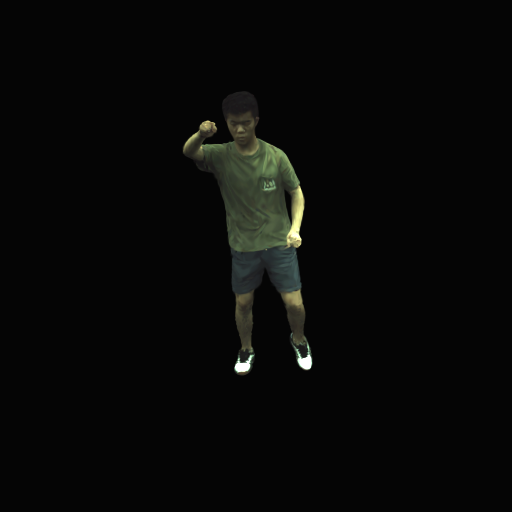}
\end{subfigure}
\begin{subfigure}[b]{0.16\textwidth}
    \includegraphics [trim=150 180 150 80, clip, width=1.0\textwidth]{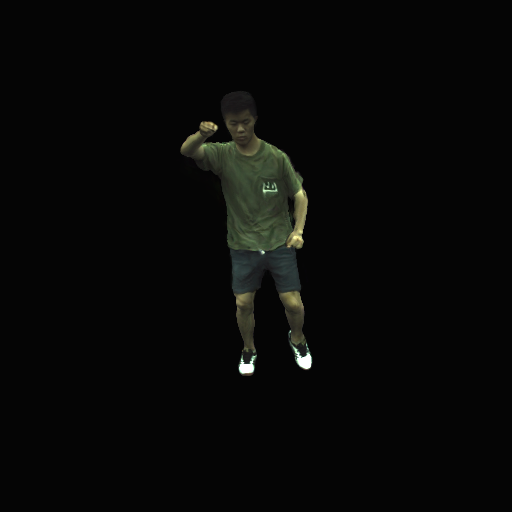}
\end{subfigure}
\begin{subfigure}[b]{0.16\textwidth}
    \includegraphics [trim=150 180 150 80, clip, width=1.0\textwidth]{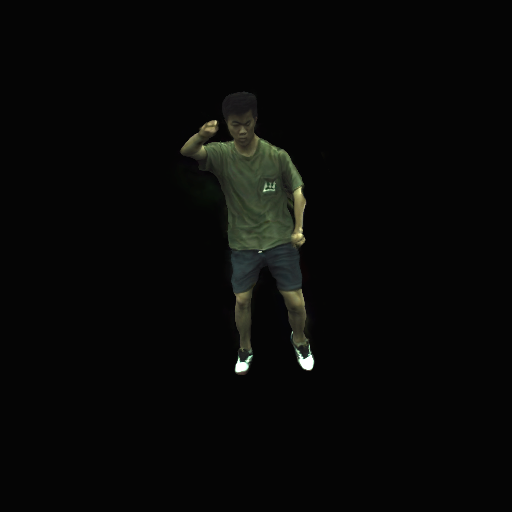}
\end{subfigure}
\begin{subfigure}[b]{0.16\textwidth}
    \includegraphics [trim=150 180 150 80, clip, width=1.0\textwidth]{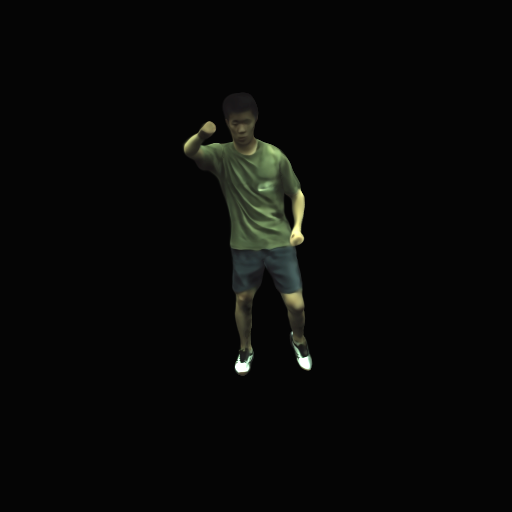}
\end{subfigure}
\begin{subfigure}[b]{0.16\textwidth}
    \includegraphics [trim=150 180 150 80, clip, width=1.0\textwidth]{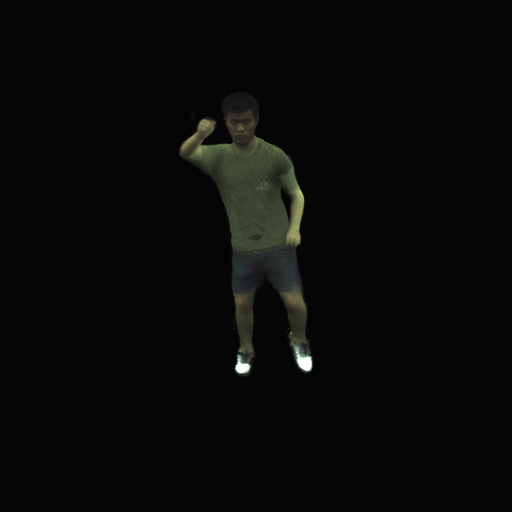}
\end{subfigure}
\\
\begin{subfigure}[b]{0.16\textwidth}
    \includegraphics [trim=160 150 140 80, clip, width=1.0\textwidth]{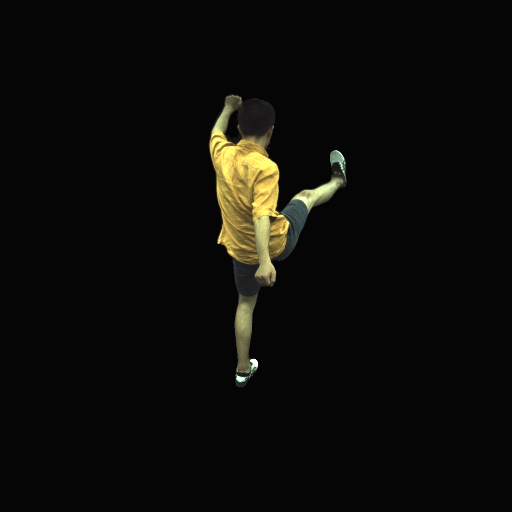}
\end{subfigure}
\begin{subfigure}[b]{0.16\textwidth}
    \includegraphics [trim=160 150 140 80, clip, width=1.0\textwidth]{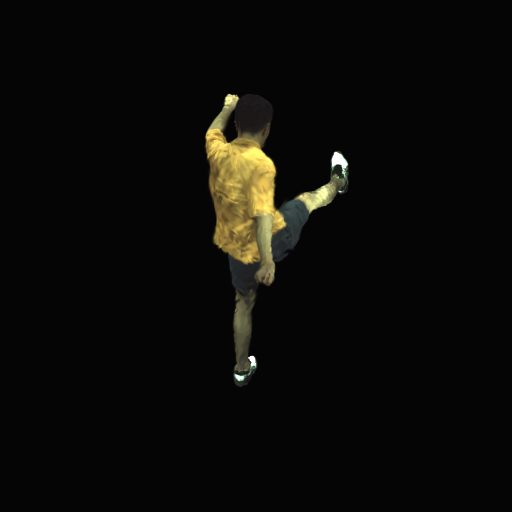}
\end{subfigure}
\begin{subfigure}[b]{0.16\textwidth}
    \includegraphics [trim=160 150 140 80, clip, width=1.0\textwidth]{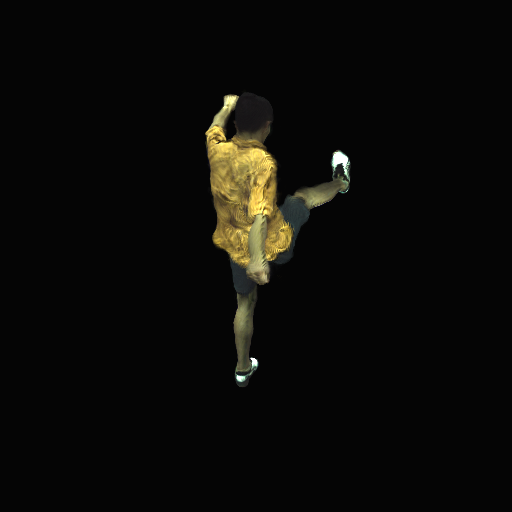}
\end{subfigure}
\begin{subfigure}[b]{0.16\textwidth}
    \includegraphics [trim=160 150 140 80, clip, width=1.0\textwidth]{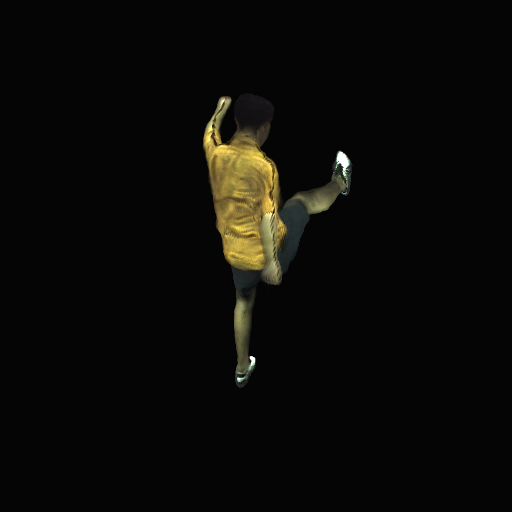}
\end{subfigure}
\begin{subfigure}[b]{0.16\textwidth}
    \includegraphics [trim=160 150 140 80, clip, width=1.0\textwidth]{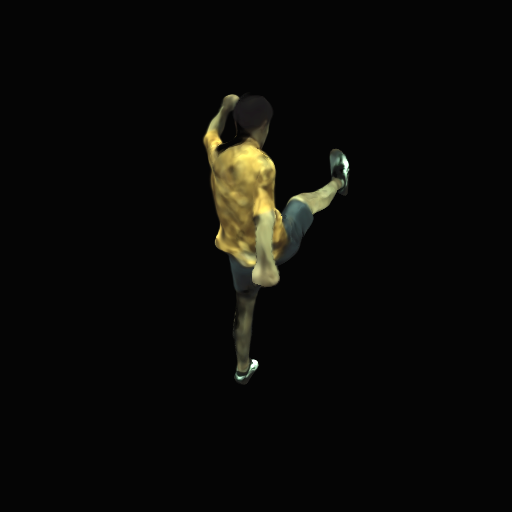}
\end{subfigure}
\begin{subfigure}[b]{0.16\textwidth}
    \includegraphics [trim=160 150 140 80, clip, width=1.0\textwidth]{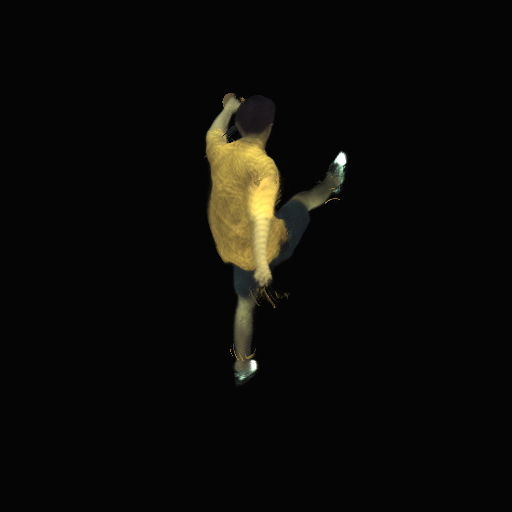}
\end{subfigure}
\\
\begin{subfigure}[b]{0.16\textwidth}
    \includegraphics [trim=120 40 120 120, clip, width=1.0\textwidth]{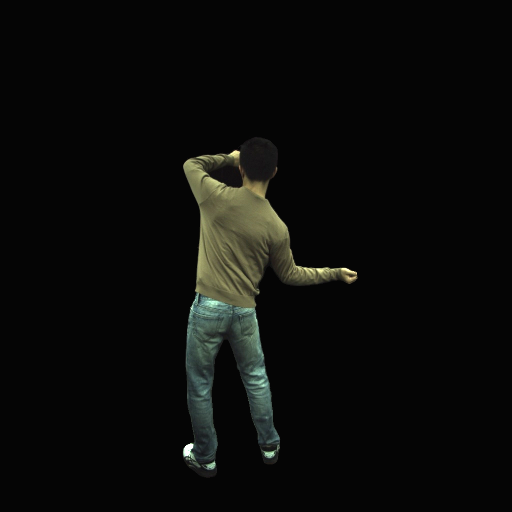}
    \caption{GT}
\end{subfigure}
\begin{subfigure}[b]{0.16\textwidth}
    \includegraphics [trim=120 40 120 120, clip, width=1.0\textwidth]{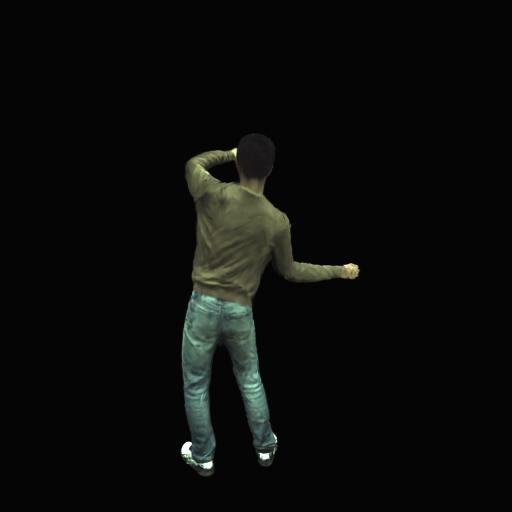}
    \caption{Ours}
\end{subfigure}
\begin{subfigure}[b]{0.16\textwidth}
    \includegraphics [trim=120 40 120 120, clip, width=1.0\textwidth]{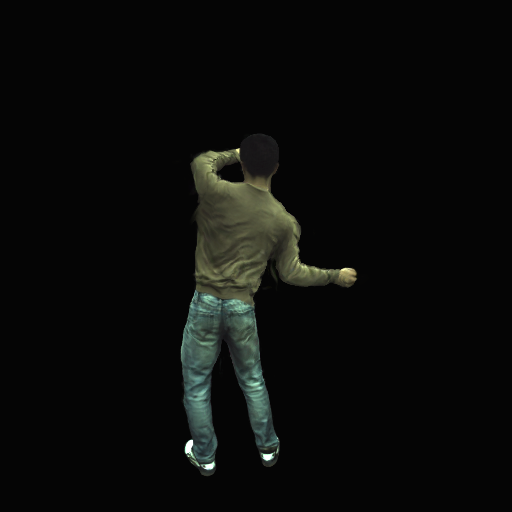}
    \caption{HumanNeRF~\cite{weng2022humannerf}}
\end{subfigure}
\begin{subfigure}[b]{0.16\textwidth}
    \includegraphics [trim=120 40 120 120, clip, width=1.0\textwidth]{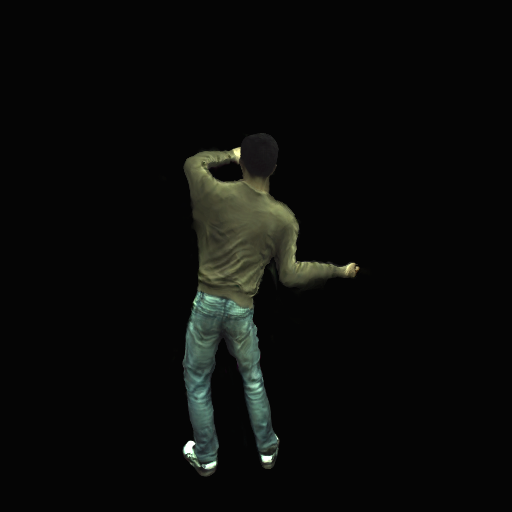}
    \caption{MonoHuman~\cite{yu2023monohuman}}
\end{subfigure}
\begin{subfigure}[b]{0.16\textwidth}
    \includegraphics [trim=120 40 120 120, clip, width=1.0\textwidth]{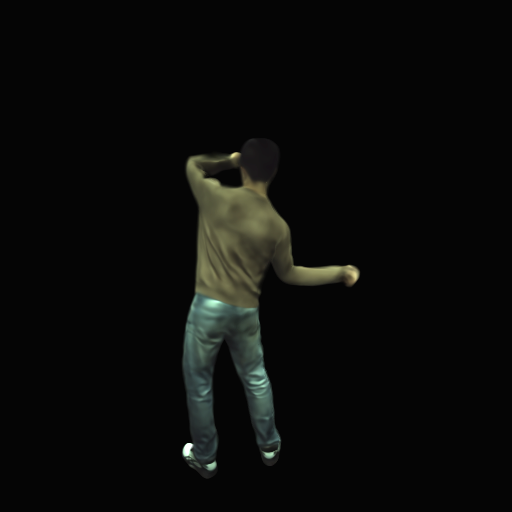}
    \caption{ARAH~\cite{ARAH:ECCV:2022}}
\end{subfigure}
\begin{subfigure}[b]{0.16\textwidth}
    \includegraphics [trim=120 40 120 120, clip, width=1.0\textwidth]{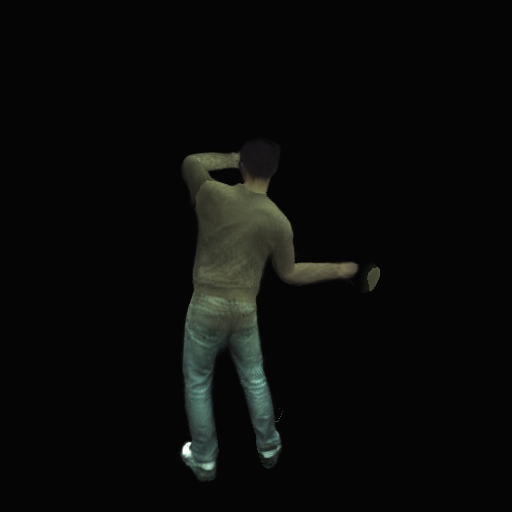}
    \caption{Instant-NVR~\cite{instant_nvr}}
\end{subfigure}
\caption{\textbf{Qualitative Comparison of Novel View Synthesis on ZJU-MoCap.} 
}
\label{fig:supp_quality_zju_nv}
\end{figure}

\subsection{Qualitative Results of Novel View Synthesis on ZJU-MoCap}

Additional qualitative comparison of novel view synthesis on ZJU-MoCap is shown in \cref{fig:supp_quality_zju_nv}. HumanNeRF and MonoHuman preserves sharp details, but often produces undesired distortions and cloud-like effect around the contour. ARAH gives more rigid body thanks to their explicit modeling of geometry, while they show misalignment and lack fine details. Instant-NVR synthesizes blurry appearance and obvious artifacts on the limbs. Overall, our method can generate high-quality images with realistic cloth deformations.

\begin{figure}[t!]
\captionsetup[subfigure]{labelformat=empty}
\scriptsize
\centering
\begin{subfigure}[b]{0.23\textwidth}
    \includegraphics [trim=140 80 50 80, clip, width=1.0\textwidth]{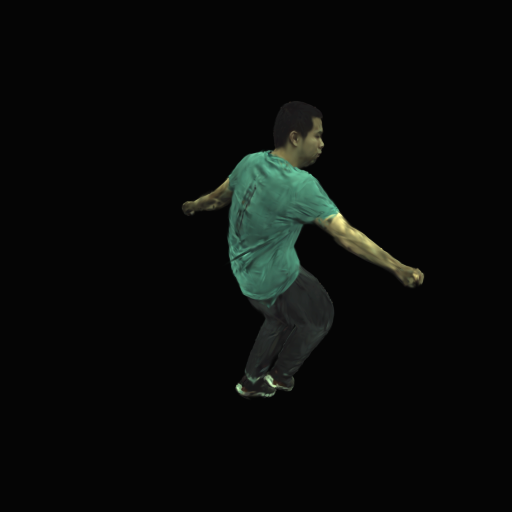}
\end{subfigure}
\begin{subfigure}[b]{0.23\textwidth}
    \includegraphics [trim=140 80 50 80, clip, width=1.0\textwidth]{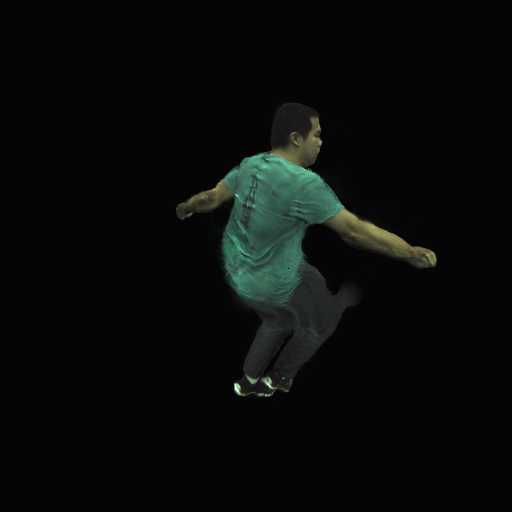}
\end{subfigure}
\begin{subfigure}[b]{0.23\textwidth}
    \includegraphics [trim=140 80 50 80, clip, width=1.0\textwidth]{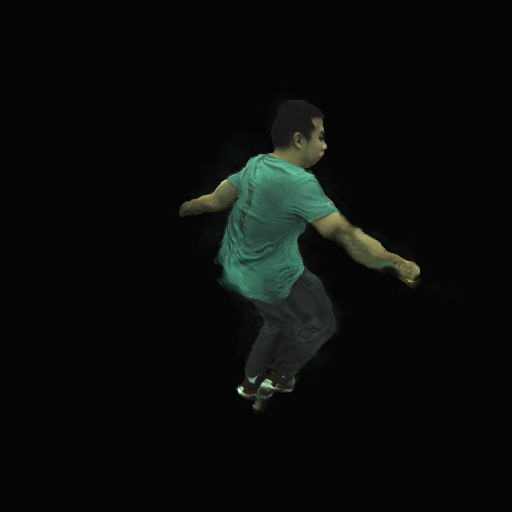}
\end{subfigure}
\begin{subfigure}[b]{0.23\textwidth}
    \includegraphics [trim=140 80 50 80, clip, width=1.0\textwidth]{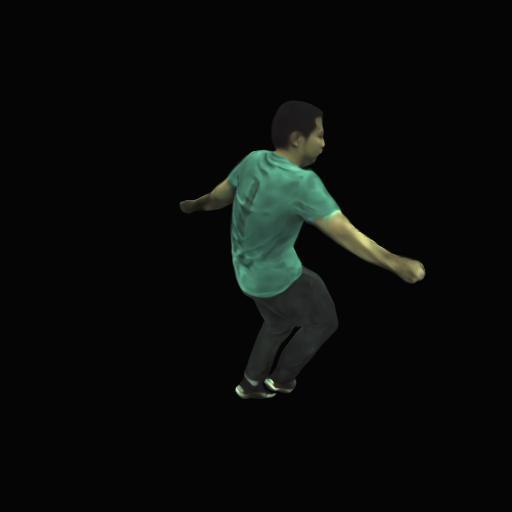}
\end{subfigure}
\\
\begin{subfigure}[b]{0.23\textwidth}
    \includegraphics [trim=100 40 150 50, clip, width=1.0\textwidth]{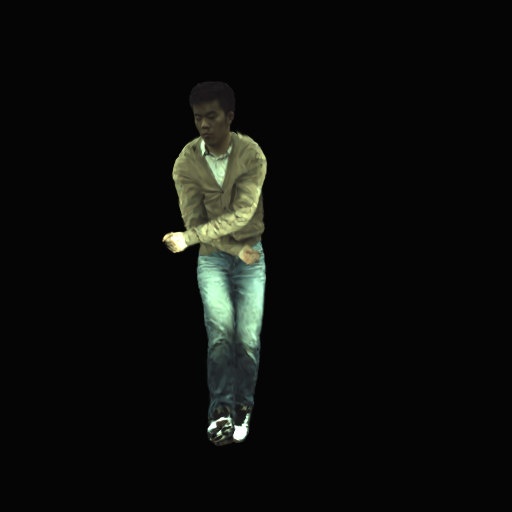}
    \caption{Ours}
\end{subfigure}
\begin{subfigure}[b]{0.23\textwidth}
    \includegraphics [trim=100 40 150 50, clip, width=1.0\textwidth]{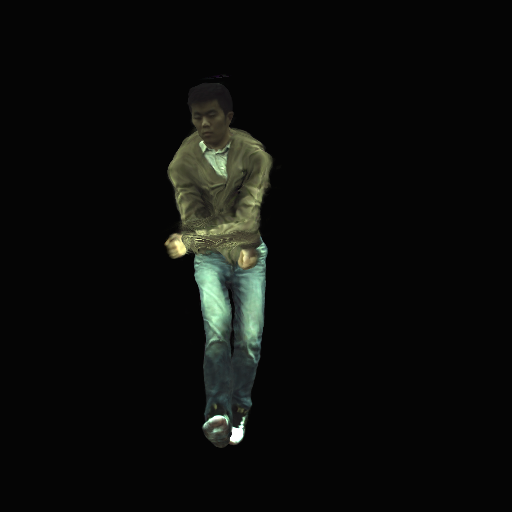}
    \caption{HumanNeRF~\cite{weng2022humannerf}}
\end{subfigure}
\begin{subfigure}[b]{0.23\textwidth}
    \includegraphics [trim=100 40 150 50, clip, width=1.0\textwidth]{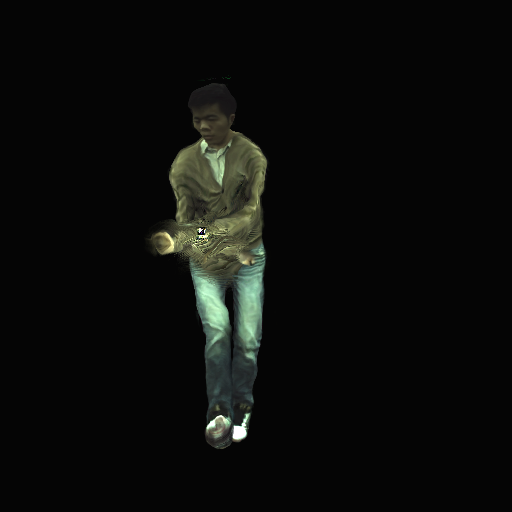}
    \caption{MonoHuman~\cite{yu2023monohuman}}
\end{subfigure}
\begin{subfigure}[b]{0.23\textwidth}
    \includegraphics [trim=100 40 150 50, clip, width=1.0\textwidth]{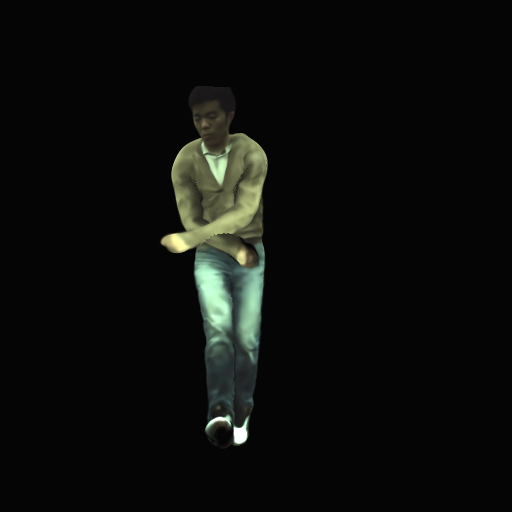}
    \caption{ARAH~\cite{ARAH:ECCV:2022}}
\end{subfigure}
\caption{\textbf{Qualitative Comparison of Out-of-distribution Pose Animation on ZJU-MoCap.} 
}
\label{fig:supp_quality_zju_odp}
\end{figure}

\subsection{Qualitative Results of Out-of-distribution Poses on ZJU-MoCap}

We present qualitative comparison of extreme out-of-distribution pose animation in \cref{fig:supp_quality_zju_odp}. Our method does not produce obvious artifacts compared to baselines, demonstrating good generalization to unseen poses.

\begin{figure}[t!]
\captionsetup[subfigure]{labelformat=empty}
\scriptsize
\centering
\begin{subfigure}[b]{0.16\textwidth}
    \includegraphics [trim=110 10 120 70, clip, width=1.0\textwidth]{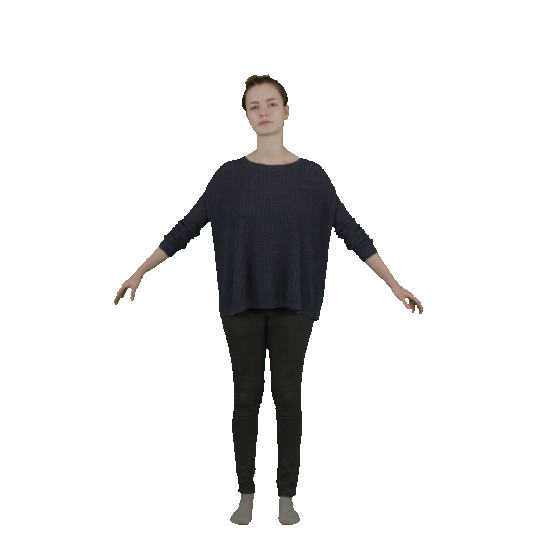}
\end{subfigure}
\begin{subfigure}[b]{0.16\textwidth}
    \includegraphics [trim=110 10 120 70, clip, width=1.0\textwidth]{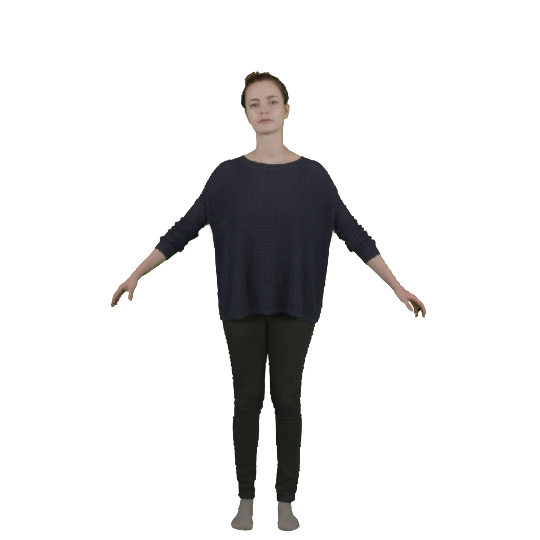}
\end{subfigure}
\begin{subfigure}[b]{0.16\textwidth}
    \includegraphics [trim=110 10 120 70, clip, width=1.0\textwidth]{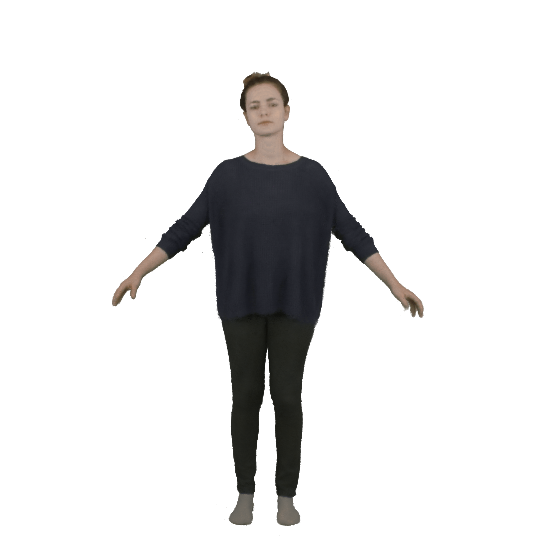}
\end{subfigure}
\begin{subfigure}[b]{0.16\textwidth}
    \includegraphics [trim=110 10 120 80, clip, width=1.0\textwidth]{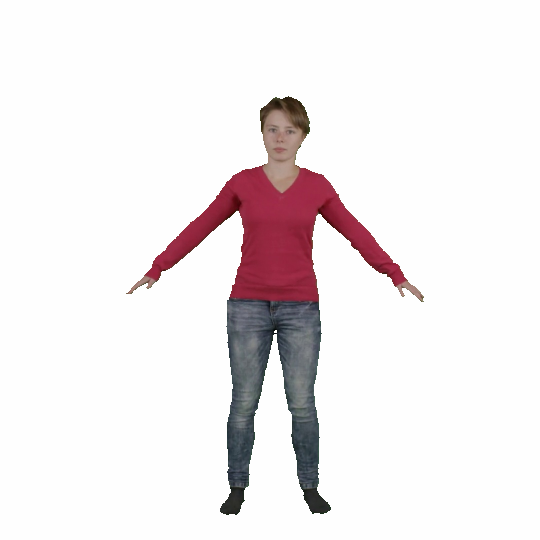}
\end{subfigure}
\begin{subfigure}[b]{0.16\textwidth}
    \includegraphics [trim=110 10 120 80, clip, width=1.0\textwidth]{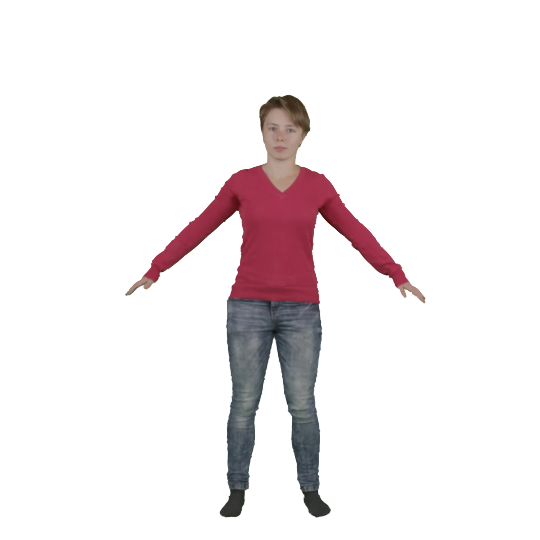}
\end{subfigure}
\begin{subfigure}[b]{0.16\textwidth}
    \includegraphics [trim=110 10 120 80, clip, width=1.0\textwidth]{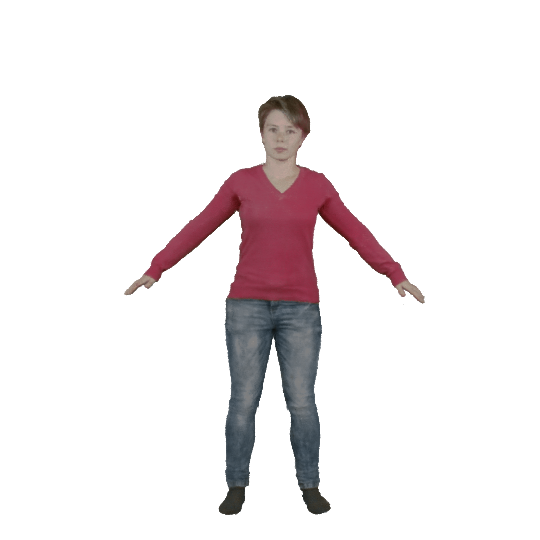}
\end{subfigure}
\\
\begin{subfigure}[b]{0.16\textwidth}
    \includegraphics [trim=130 10 100 70, clip, width=1.0\textwidth]{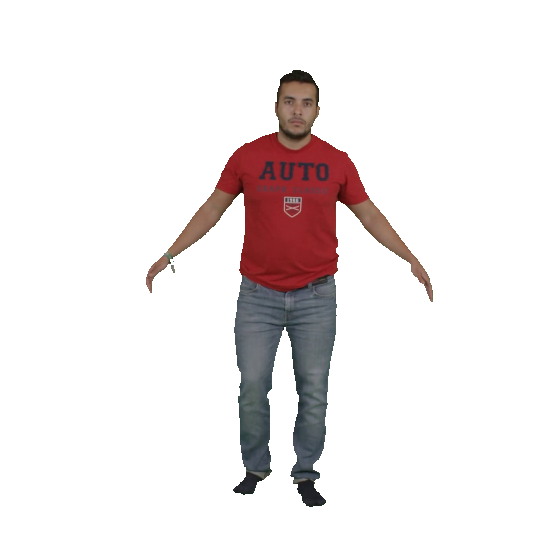}
    \caption{GT}
\end{subfigure}
\begin{subfigure}[b]{0.16\textwidth}
    \includegraphics [trim=130 10 100 70, clip, width=1.0\textwidth]{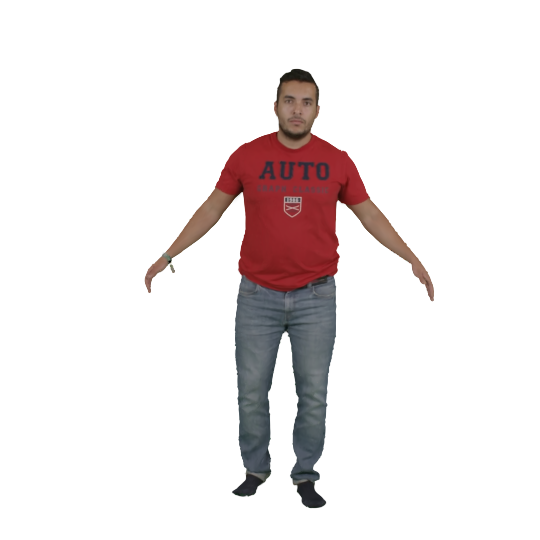}
    \caption{Ours}
\end{subfigure}
\begin{subfigure}[b]{0.16\textwidth}
    \includegraphics [trim=130 10 100 70, clip, width=1.0\textwidth]{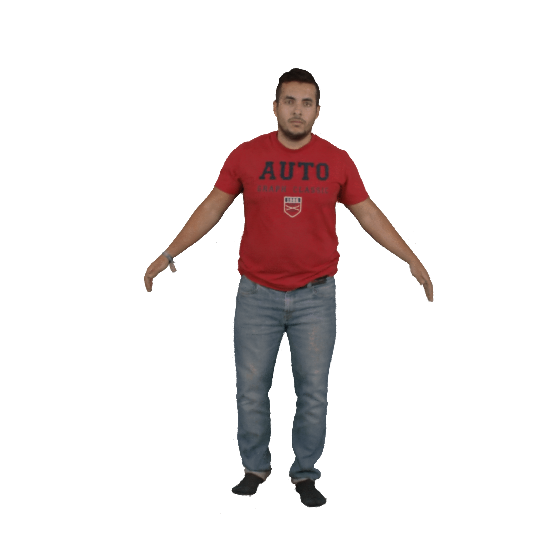}
    \caption{InstantAvatar~\cite{jiang2022instantavatar}}
\end{subfigure}
\begin{subfigure}[b]{0.16\textwidth}
    \includegraphics [trim=120 10 100 40, clip, width=1.0\textwidth]{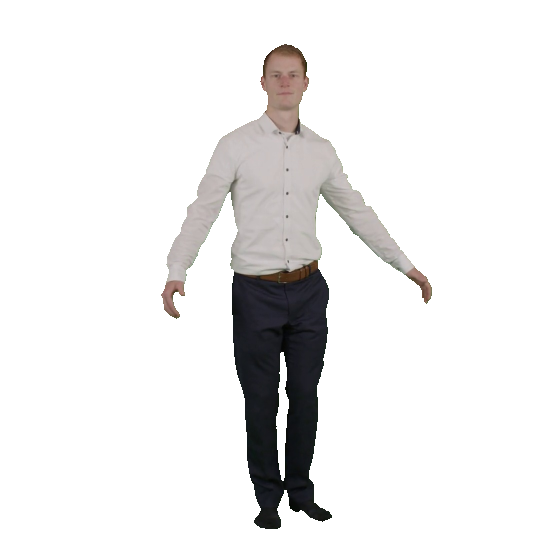}
    \caption{GT}
\end{subfigure}
\begin{subfigure}[b]{0.16\textwidth}
    \includegraphics [trim=120 10 100 40, clip, width=1.0\textwidth]{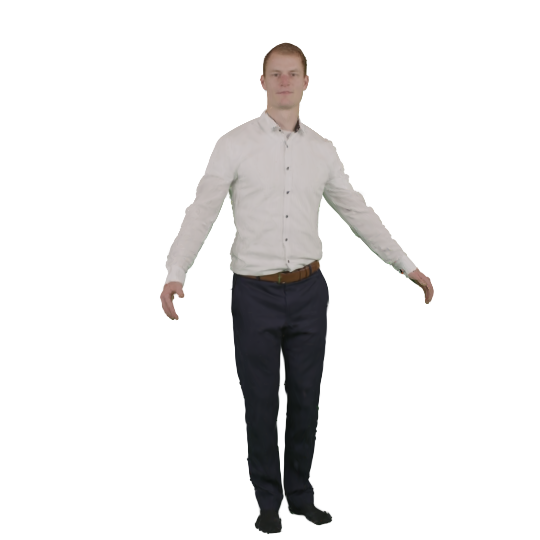}
    \caption{Ours}
\end{subfigure}
\begin{subfigure}[b]{0.16\textwidth}
    \includegraphics [trim=120 10 100 40, clip, width=1.0\textwidth]{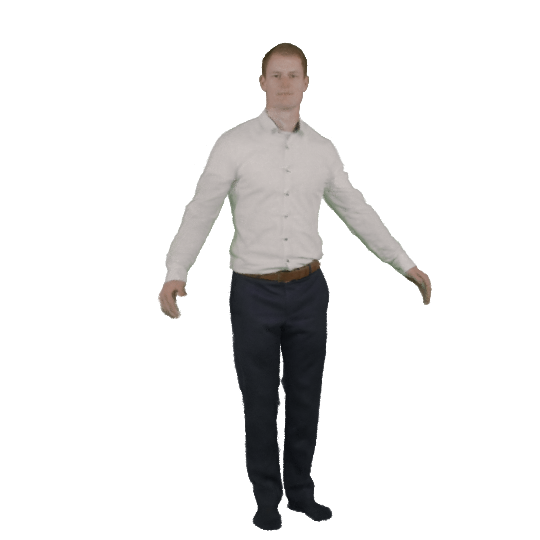}
    \caption{InstantAvatar~\cite{jiang2022instantavatar}}
\end{subfigure}
\caption{\textbf{Qualitative Comparison on PeopleSnapshot~\cite{alldieck2018video}. Best viewed zoomed-in.} 
}
\label{fig:supp_quality_ps}
\end{figure}

\subsection{Qualitative Results on PeopleSnapshot}

We show qualitative results on the test set of PeopleSnapshot in \cref{fig:supp_quality_ps}. Compared to InstantAvatar, our method produces sharper results, especially in the face region.

%% file: sec/supplement/limitation.tex
\section{Limitations}
\label{sec:supp_limitations}


While our proposed approach achieves state-of-the-art rendering quality of clothed human avatars with an interactive frame rate of rendering, the training time of our model still does not match those fast grid-based methods~\cite{jiang2022instantavatar,instant_nvr}. On the other hand, our method may produce blurry results in areas with high-frequency texture or repetitive patterns, such as striped shirts. Lastly, our method does not provide accurate geometry reconstruction of the avatar, unlike ARAH~\cite{ARAH:ECCV:2022}. 
Despite reasonable rendering quality, our method generates noisy surface normal resulting from the inconsistency of Gaussian splat depth. It would be particularly interesting to study how to extract a smooth, detailed geometry from the 3DGS avatar model, possibly by applying regularization to the normal map or attaching 3D Gaussians to an underlying mesh. 

%% file: main.bbl
\begin{thebibliography}{78}
\providecommand{\natexlab}[1]{#1}
\providecommand{\url}[1]{\texttt{#1}}
\expandafter\ifx\csname urlstyle\endcsname\relax
  \providecommand{\doi}[1]{doi: #1}\else
  \providecommand{\doi}{doi: \begingroup \urlstyle{rm}\Url}\fi

\bibitem[Alldieck et~al.(2018)Alldieck, Magnor, Xu, Theobalt, and
  Pons-Moll]{alldieck2018video}
Thiemo Alldieck, Marcus Magnor, Weipeng Xu, Christian Theobalt, and Gerard
  Pons-Moll.
\newblock Video based reconstruction of 3d people models.
\newblock In \emph{Proc. of CVPR}, 2018.

\bibitem[Anguelov et~al.(2005)Anguelov, Srinivasan, Koller, Thrun, Rodgers, and
  Davis]{SCAPE}
Dragomir Anguelov, Praveen Srinivasan, Daphne Koller, Sebastian Thrun, Jim
  Rodgers, and James Davis.
\newblock Scape: shape completion and animation of people.
\newblock \emph{ACM Transasctions Graphics}, 24, 2005.

\bibitem[Chen et~al.(2022)Chen, Xu, Geiger, Yu, and Su]{TensoRF:ECCV:2022}
Anpei Chen, Zexiang Xu, Andreas Geiger, Jingyi Yu, and Hao Su.
\newblock Tensorf: Tensorial radiance fields.
\newblock In \emph{Proc. of ECCV}, 2022.

\bibitem[Chen et~al.(2021)Chen, Zheng, Black, Hilliges, and
  Geiger]{Chen2021ICCV}
Xu Chen, Yufeng Zheng, Michael Black, Otmar Hilliges, and Andreas Geiger.
\newblock Snarf: Differentiable forward skinning for animating non-rigid neural
  implicit shapes.
\newblock In \emph{Proc. of ICCV}, 2021.

\bibitem[Chen et~al.(2023{\natexlab{a}})Chen, Jiang, Song, Rietmann, Geiger,
  Black, and Hilliges]{Chen2023PAMI}
Xu Chen, Tianjian Jiang, Jie Song, Max Rietmann, Andreas Geiger, Michael~J.
  Black, and Otmar Hilliges.
\newblock Fast-snarf: A fast deformer for articulated neural fields.
\newblock \emph{Pattern Analysis and Machine Intelligence (PAMI)},
  2023{\natexlab{a}}.

\bibitem[Chen et~al.(2023{\natexlab{b}})Chen, Wang, Chen, Zhang, Li, Guo, Wang,
  and Wang]{chen2023uv}
Yue Chen, Xuan Wang, Xingyu Chen, Qi Zhang, Xiaoyu Li, Yu Guo, Jue Wang, and
  Fei Wang.
\newblock Uv volumes for real-time rendering of editable free-view human
  performance.
\newblock In \emph{Proc. of CVPR}, 2023{\natexlab{b}}.

\bibitem[Geng et~al.(2023)Geng, Peng, Xu, Bao, and Zhou]{instant_nvr}
Chen Geng, Sida Peng, Zhen Xu, Hujun Bao, and Xiaowei Zhou.
\newblock Learning neural volumetric representations of dynamic humans in
  minutes.
\newblock In \emph{Proc. of CVPR}, 2023.

\bibitem[Guo et~al.(2023)Guo, Jiang, Chen, Song, and
  Hilliges]{guo2023vid2avatar}
Chen Guo, Tianjian Jiang, Xu Chen, Jie Song, and Otmar Hilliges.
\newblock Vid2avatar: 3d avatar reconstruction from videos in the wild via
  self-supervised scene decomposition.
\newblock In \emph{Proc. of CVPR}, 2023.

\bibitem[Hasler et~al.(2009)Hasler, Stoll, Sunkel, Rosenhahn, and
  Seidel]{Hasler2009CGF}
N. Hasler, C. Stoll, M. Sunkel, B. Rosenhahn, and H.-P. Seidel.
\newblock {A Statistical Model of Human Pose and Body Shape}.
\newblock \emph{Computer Graphics Forum}, 28:\penalty0 337--346, 2009.

\bibitem[Hu and Liu(2024)]{hu2023gauhuman}
Shoukang Hu and Ziwei Liu.
\newblock Gauhuman: Articulated gaussian splatting from monocular human videos.
\newblock In \emph{Proc. of CVPR}, 2024.

\bibitem[Jena et~al.(2023)Jena, Iyer, Choudhary, Smith, Chaudhari, and
  Gee]{jena2023splatarmor}
Rohit Jena, Ganesh~Subramanian Iyer, Siddharth Choudhary, Brandon Smith, Pratik
  Chaudhari, and James Gee.
\newblock Splatarmor: Articulated gaussian splatting for animatable humans from
  monocular rgb videos.
\newblock \emph{arXiv preprint arXiv:2311.10812}, 2023.

\bibitem[Jiang et~al.(2023)Jiang, Chen, Song, and
  Hilliges]{jiang2022instantavatar}
Tianjian Jiang, Xu Chen, Jie Song, and Otmar Hilliges.
\newblock Instantavatar: Learning avatars from monocular video in 60 seconds.
\newblock In \emph{Proc. of CVPR}, 2023.

\bibitem[Jiang et~al.(2022)Jiang, Yi, Samei, Tuzel, and
  Ranjan]{jiang2022neuman}
Wei Jiang, Kwang~Moo Yi, Golnoosh Samei, Oncel Tuzel, and Anurag Ranjan.
\newblock Neuman: Neural human radiance field from a single video.
\newblock In \emph{Proc. of ECCV}, 2022.

\bibitem[Kerbl et~al.(2023)Kerbl, Kopanas, Leimk{\"u}hler, and
  Drettakis]{kerbl3Dgaussians}
Bernhard Kerbl, Georgios Kopanas, Thomas Leimk{\"u}hler, and George Drettakis.
\newblock 3d gaussian splatting for real-time radiance field rendering.
\newblock \emph{ACM Transactions on Graphics}, 42\penalty0 (4), 2023.

\bibitem[Kilian et~al.(2007)Kilian, Mitra, and
  Pottmann]{kmp_shape_space_sig_07}
Martin Kilian, Niloy~J. Mitra, and Helmut Pottmann.
\newblock Geometric modeling in shape space.
\newblock \emph{ACM Transactions on Graphics (SIGGRAPH)}, 26\penalty0 (3),
  2007.

\bibitem[Kingma and Ba(2015)]{Adam:ICLR:2015}
Diederik~P. Kingma and Jimmy Ba.
\newblock Adam: A method for stochastic optimization.
\newblock In \emph{Proc. of ICLR}, 2015.

\bibitem[Kocabas et~al.(2024)Kocabas, Chang, Gabriel, Tuzel, and
  Ranjan]{kocabas2023hugs}
Muhammed Kocabas, Jen-Hao~Rick Chang, James Gabriel, Oncel Tuzel, and Anurag
  Ranjan.
\newblock Hugs: Human gaussian splatting.
\newblock In \emph{Proc. of CVPR}, 2024.

\bibitem[Kwon et~al.(2023)Kwon, Liu, Fuchs, Habermann, and
  Theobalt]{kwon2023deliffas}
Youngjoong Kwon, Lingjie Liu, Henry Fuchs, Marc Habermann, and Christian
  Theobalt.
\newblock Deliffas: Deformable light fields for fast avatar synthesis.
\newblock \emph{Proc. of NeurIPS}, 2023.

\bibitem[Lei et~al.(2024)Lei, Wang, Pavlakos, Liu, and Daniilidis]{lei2023gart}
Jiahui Lei, Yufu Wang, Georgios Pavlakos, Lingjie Liu, and Kostas Daniilidis.
\newblock Gart: Gaussian articulated template models.
\newblock In \emph{Proc. of CVPR}, 2024.

\bibitem[Li et~al.(2021)Li, Yang, Ross, and Kanazawa]{aist++:ICCV:2021}
Ruilong Li, Shan Yang, David~A. Ross, and Angjoo Kanazawa.
\newblock Ai choreographer: Music conditioned 3d dance generation with aist++.
\newblock In \emph{Proc. of ICCV}, 2021.

\bibitem[Li et~al.(2022)Li, Tanke, Vo, Zollhoefer, Gall, Kanazawa, and
  Lassner]{li2022tava}
Ruilong Li, Julian Tanke, Minh Vo, Michael Zollhoefer, Jürgen Gall, Angjoo
  Kanazawa, and Christoph Lassner.
\newblock Tava: Template-free animatable volumetric actors.
\newblock In \emph{Proc. of ECCV}, 2022.

\bibitem[Li et~al.(2023)Li, Gao, Tancik, and Kanazawa]{NeRFAcc:ICCV:2023}
Ruilong Li, Hang Gao, Matthew Tancik, and Angjoo Kanazawa.
\newblock Nerfacc: Efficient sampling accelerates nerfs.
\newblock \emph{Proc. of ICCV}, 2023.

\bibitem[Li et~al.(2024)Li, Zheng, Wang, and Liu]{li2023animatable}
Zhe Li, Zerong Zheng, Lizhen Wang, and Yebin Liu.
\newblock Animatable gaussians: Learning pose-dependent gaussian maps for
  high-fidelity human avatar modeling.
\newblock In \emph{Proc. of CVPR}, 2024.

\bibitem[Liu et~al.(2021)Liu, Habermann, Rudnev, Sarkar, Gu, and
  Theobalt]{liu2021neural}
Lingjie Liu, Marc Habermann, Viktor Rudnev, Kripasindhu Sarkar, Jiatao Gu, and
  Christian Theobalt.
\newblock Neural actor: Neural free-view synthesis of human actors with pose
  control.
\newblock \emph{ACM Trans. Graph.(ACM SIGGRAPH Asia)}, 2021.

\bibitem[Liu et~al.(2023)Liu, Huang, Qin, Lin, and Wang]{liu2023animatable}
Yang Liu, Xiang Huang, Minghan Qin, Qinwei Lin, and Haoqian Wang.
\newblock Animatable 3d gaussian: Fast and high-quality reconstruction of
  multiple human avatars, 2023.

\bibitem[Loper et~al.(2015)Loper, Mahmood, Romero, Pons-Moll, and
  Black]{SMPL:2015}
Matthew Loper, Naureen Mahmood, Javier Romero, Gerard Pons-Moll, and Michael~J.
  Black.
\newblock {SMPL}: A skinned multi-person linear model.
\newblock \emph{ACM Transasctions Graphics}, 34\penalty0 (6), 2015.

\bibitem[Mahmood et~al.(2019)Mahmood, Ghorbani, Troje, Pons-Moll, and
  Black]{AMASS:ICCV:2019}
Naureen Mahmood, Nima Ghorbani, Nikolaus~F. Troje, Gerard Pons-Moll, and
  Michael~J. Black.
\newblock {AMASS}: Archive of motion capture as surface shapes.
\newblock In \emph{Proc. of ICCV}, 2019.

\bibitem[Mihajlovic et~al.(2021)Mihajlovic, Zhang, Black, and
  Tang]{LEAP:CVPR:21}
Marko Mihajlovic, Yan Zhang, Michael~J. Black, and Siyu Tang.
\newblock {LEAP}: Learning articulated occupancy of people.
\newblock In \emph{Proc. of CVPR}, 2021.

\bibitem[Mildenhall et~al.(2020)Mildenhall, Srinivasan, Tancik, Barron,
  Ramamoorthi, and Ng]{mildenhall2020nerf}
Ben Mildenhall, Pratul~P. Srinivasan, Matthew Tancik, Jonathan~T. Barron, Ravi
  Ramamoorthi, and Ren Ng.
\newblock Nerf: Representing scenes as neural radiance fields for view
  synthesis.
\newblock In \emph{Proc. of ECCV}, 2020.

\bibitem[Moreau et~al.(2024)Moreau, Song, Dhamo, Shaw, Zhou, and
  Pérez-Pellitero]{moreau2023human}
Arthur Moreau, Jifei Song, Helisa Dhamo, Richard Shaw, Yiren Zhou, and Eduardo
  Pérez-Pellitero.
\newblock Human gaussian splatting: Real-time rendering of animatable avatars.
\newblock In \emph{Proc. of CVPR}, 2024.

\bibitem[M\"uller et~al.(2022)M\"uller, Evans, Schied, and
  Keller]{mueller2022instant}
Thomas M\"uller, Alex Evans, Christoph Schied, and Alexander Keller.
\newblock Instant neural graphics primitives with a multiresolution hash
  encoding.
\newblock \emph{ACM Transasctions Graphics}, 41\penalty0 (4), 2022.

\bibitem[Niemeyer et~al.(2020)Niemeyer, Mescheder, Oechsle, and
  Geiger]{Niemeyer2020CVPR}
Michael Niemeyer, Lars Mescheder, Michael Oechsle, and Andreas Geiger.
\newblock Differentiable volumetric rendering: Learning implicit 3d
  representations without 3d supervision.
\newblock In \emph{Proc. of CVPR}, 2020.

\bibitem[Noguchi et~al.(2021)Noguchi, Sun, Lin, and Harada]{NARF:ICCV:2021}
Atsuhiro Noguchi, Xiao Sun, Stephen Lin, and Tatsuya Harada.
\newblock Neural articulated radiance field.
\newblock In \emph{Proc. of ICCV}, 2021.

\bibitem[Oechsle et~al.(2021)Oechsle, Peng, and Geiger]{Oechsle2021ICCV}
Michael Oechsle, Songyou Peng, and Andreas Geiger.
\newblock Unisurf: Unifying neural implicit surfaces and radiance fields for
  multi-view reconstruction.
\newblock In \emph{Proc. of ICCV}, 2021.

\bibitem[Osman et~al.(2020)Osman, Bolkart, and Black]{STAR:ECCV:2020}
Ahmed A.~A. Osman, Timo Bolkart, and Michael~J. Black.
\newblock Star: Sparse trained articulated human body regressor.
\newblock In \emph{Proc. of ECCV}, 2020.

\bibitem[Pavlakos et~al.(2018)Pavlakos, Zhu, Zhou, and
  Daniilidis]{Pavlakos_2018_CVPR}
Georgios Pavlakos, Luyang Zhu, Xiaowei Zhou, and Kostas Daniilidis.
\newblock Learning to estimate 3d human pose and shape from a single color
  image.
\newblock In \emph{Proc. of CVPR}, 2018.

\bibitem[Peng et~al.(2021{\natexlab{a}})Peng, Dong, Wang, Zhang, Shuai, Zhou,
  and Bao]{peng2021animatable}
Sida Peng, Junting Dong, Qianqian Wang, Shangzhan Zhang, Qing Shuai, Xiaowei
  Zhou, and Hujun Bao.
\newblock Animatable neural radiance fields for modeling dynamic human bodies.
\newblock In \emph{Proc. of ICCV}, 2021{\natexlab{a}}.

\bibitem[Peng et~al.(2021{\natexlab{b}})Peng, Jiang, Liao, Niemeyer, Pollefeys,
  and Geiger]{Peng2021SAP}
Songyou Peng, Chiyu~"Max" Jiang, Yiyi Liao, Michael Niemeyer, Marc Pollefeys,
  and Andreas Geiger.
\newblock Shape as points: A differentiable poisson solver.
\newblock In \emph{Proc. of NeurIPS}, 2021{\natexlab{b}}.

\bibitem[Peng et~al.(2021{\natexlab{c}})Peng, Zhang, Xu, Wang, Shuai, Bao, and
  Zhou]{peng2020neural}
Sida Peng, Yuanqing Zhang, Yinghao Xu, Qianqian Wang, Qing Shuai, Hujun Bao,
  and Xiaowei Zhou.
\newblock Neural body: Implicit neural representations with structured latent
  codes for novel view synthesis of dynamic humans.
\newblock In \emph{Proc. of CVPR}, 2021{\natexlab{c}}.

\bibitem[Peng et~al.(2022)Peng, Zhang, Xu, Geng, Jiang, Bao, and
  Zhou]{peng2022animatable}
Sida Peng, Shangzhan Zhang, Zhen Xu, Chen Geng, Boyi Jiang, Hujun Bao, and
  Xiaowei Zhou.
\newblock Animatable neural implicit surfaces for creating avatars from videos.
\newblock \emph{ArXiv}, abs/2203.08133, 2022.

\bibitem[Peng et~al.(2023)Peng, Yan, Shuai, Bao, and Zhou]{peng2023mlpmaps}
Sida Peng, Yunzhi Yan, Qing Shuai, Hujun Bao, and Xiaowei Zhou.
\newblock Representing volumetric videos as dynamic mlp maps.
\newblock In \emph{Proc. of CVPR}, 2023.

\bibitem[Prokudin et~al.(2021)Prokudin, Black, and Romero]{SMPLpix:WACV:2020}
Sergey Prokudin, Michael~J. Black, and Javier Romero.
\newblock {SMPLpix}: Neural avatars from {3D} human models.
\newblock In \emph{Proc. of WACV}, 2021.

\bibitem[Prokudin et~al.(2023)Prokudin, Ma, Raafat, Valentin, and
  Tang]{Prokudin_2023_ICCV}
Sergey Prokudin, Qianli Ma, Maxime Raafat, Julien Valentin, and Siyu Tang.
\newblock Dynamic point fields.
\newblock In \emph{Proc. of ICCV}, 2023.

\bibitem[Raj et~al.(2021)Raj, Tanke, Hays, Vo, Stoll, and Lassner]{raj2020anr}
Amit Raj, Julian Tanke, James Hays, Minh Vo, Carsten Stoll, and Christoph
  Lassner.
\newblock Anr-articulated neural rendering for virtual avatars.
\newblock In \emph{Proc. of CVPR}, 2021.

\bibitem[Reiser et~al.(2023)Reiser, Szeliski, Verbin, Srinivasan, Mildenhall,
  Geiger, Barron, and Hedman]{Reiser2023SIGGRAPH}
Christian Reiser, Richard Szeliski, Dor Verbin, Pratul~P. Srinivasan, Ben
  Mildenhall, Andreas Geiger, Jonathan~T. Barron, and Peter Hedman.
\newblock Merf: Memory-efficient radiance fields for real-time view synthesis
  in unbounded scenes.
\newblock \emph{ACM TOG}, 42\penalty0 (4), 2023.

\bibitem[R\"{u}ckert et~al.(2022)R\"{u}ckert, Franke, and
  Stamminger]{ruckert2021adop}
Darius R\"{u}ckert, Linus Franke, and Marc Stamminger.
\newblock Adop: Approximate differentiable one-pixel point rendering.
\newblock \emph{ACM Transactions on Graphics}, 41\penalty0 (4), 2022.

\bibitem[Saito et~al.(2021)Saito, Yang, Ma, and Black]{SCANimate:CVPR:21}
Shunsuke Saito, Jinlong Yang, Qianli Ma, and Michael~J. Black.
\newblock {SCANimate}: Weakly supervised learning of skinned clothed avatar
  networks.
\newblock In \emph{Proc. of CVPR}, 2021.

\bibitem[{Sara Fridovich-Keil and Alex Yu} et~al.(2022){Sara Fridovich-Keil and
  Alex Yu}, Tancik, Chen, Recht, and
  Kanazawa]{yu_and_fridovichkeil2021plenoxels}
{Sara Fridovich-Keil and Alex Yu}, Matthew Tancik, Qinhong Chen, Benjamin
  Recht, and Angjoo Kanazawa.
\newblock Plenoxels: Radiance fields without neural networks.
\newblock In \emph{Proc. of CVPR}, 2022.

\bibitem[Sitzmann et~al.(2021)Sitzmann, Rezchikov, Freeman, Tenenbaum, and
  Durand]{sitzmann2021lfns}
Vincent Sitzmann, Semon Rezchikov, William~T. Freeman, Joshua~B. Tenenbaum, and
  Fredo Durand.
\newblock Light field networks: Neural scene representations with
  single-evaluation rendering.
\newblock In \emph{Proc. of NeurIPS}, 2021.

\bibitem[Su et~al.(2021)Su, Yu, Zollhoefer, and Rhodin]{ANeRF:NeurIPS:2021}
Shih-Yang Su, Frank Yu, Michael Zollhoefer, and Helge Rhodin.
\newblock A-ne{RF}: Articulated neural radiance fields for learning human
  shape, appearance, and pose.
\newblock In \emph{Proc. of NeurIPS}, 2021.

\bibitem[Su et~al.(2023)Su, Bagautdinov, and Rhodin]{su2023npc}
Shih-Yang Su, Timur Bagautdinov, and Helge Rhodin.
\newblock Npc: Neural point characters from video.
\newblock In \emph{Proc. of ICCV}, 2023.

\bibitem[Suhail et~al.(2022)Suhail, Esteves, Sigal, and
  Makadia]{suhail2022generalizable}
Mohammed Suhail, Carlos Esteves, Leonid Sigal, and Ameesh Makadia.
\newblock Generalizable patch-based neural rendering.
\newblock In \emph{Proc. of ECCV}, 2022.

\bibitem[Suhail1 et~al.(2022)Suhail1, Esteves, Sigal, and
  Makadia]{suhail2022lightfield}
Mohammed Suhail1, Carlos Esteves, Leonid Sigal, and Ameesh Makadia.
\newblock Light field neural rendering.
\newblock In \emph{Proc. of CVPR}, 2022.

\bibitem[Sun et~al.(2022)Sun, Sun, and Chen]{DVGO:CVPR:2022}
Cheng Sun, Min Sun, and Hwann{-}Tzong Chen.
\newblock Direct voxel grid optimization: Super-fast convergence for radiance
  fields reconstruction.
\newblock In \emph{Proc. of CVPR}, 2022.

\bibitem[Wang et~al.(2022{\natexlab{a}})Wang, Ren, Huang, Olszewski, Chai, Fu,
  and Tulyakov]{wang2022r2l}
Huan Wang, Jian Ren, Zeng Huang, Kyle Olszewski, Menglei Chai, Yun Fu, and
  Sergey Tulyakov.
\newblock R2l: Distilling neural radiance field to neural light field for
  efficient novel view synthesis.
\newblock In \emph{Proc. of ECCV}, 2022{\natexlab{a}}.

\bibitem[Wang et~al.(2022{\natexlab{b}})Wang, Zhang, Liu, Zhao, Zhang, Zhang,
  Wu, Yu, and Xu]{FourierOctree:CVPR:2022}
Liao Wang, Jiakai Zhang, Xinhang Liu, Fuqiang Zhao, Yanshun Zhang, Yingliang
  Zhang, Minye Wu, Jingyi Yu, and Lan Xu.
\newblock Fourier plenoctrees for dynamic radiance field rendering in
  real-time.
\newblock In \emph{Proc. of CVPR}, 2022{\natexlab{b}}.

\bibitem[Wang et~al.(2021{\natexlab{a}})Wang, Liu, Liu, Theobalt, Komura, and
  Wang]{wang2021neus}
Peng Wang, Lingjie Liu, Yuan Liu, Christian Theobalt, Taku Komura, and Wenping
  Wang.
\newblock Neus: Learning neural implicit surfaces by volume rendering for
  multi-view reconstruction.
\newblock In \emph{Proc. of NeurIPS}, 2021{\natexlab{a}}.

\bibitem[Wang et~al.(2021{\natexlab{b}})Wang, Mihajlovic, Ma, Geiger, and
  Tang]{MetaAvatar:NeurIPS:2021}
Shaofei Wang, Marko Mihajlovic, Qianli Ma, Andreas Geiger, and Siyu Tang.
\newblock Metaavatar: Learning animatable clothed human models from few depth
  images.
\newblock In \emph{Proc. of NeurIPS}, 2021{\natexlab{b}}.

\bibitem[Wang et~al.(2022{\natexlab{c}})Wang, Schwarz, Geiger, and
  Tang]{ARAH:ECCV:2022}
Shaofei Wang, Katja Schwarz, Andreas Geiger, and Siyu Tang.
\newblock Arah: Animatable volume rendering of articulated human sdfs.
\newblock In \emph{Proc. of ECCV}, 2022{\natexlab{c}}.

\bibitem[Weng et~al.(2022)Weng, Curless, Srinivasan, Barron, and
  Kemelmacher-Shlizerman]{weng2022humannerf}
Chung-Yi Weng, Brian Curless, Pratul~P. Srinivasan, Jonathan~T. Barron, and Ira
  Kemelmacher-Shlizerman.
\newblock Humannerf: Free-viewpoint rendering of moving people from monocular
  video.
\newblock In \emph{Proc. of CVPR}, 2022.

\bibitem[Wu et~al.(2023)Wu, Yi, Fang, Xie, Zhang, Wei, Liu, Tian, and
  Xinggang]{wu20234dgaussians}
Guanjun Wu, Taoran Yi, Jiemin Fang, Lingxi Xie, Xiaopeng Zhang, Wei Wei, Wenyu
  Liu, Qi Tian, and Wang Xinggang.
\newblock 4d gaussian splatting for real-time dynamic scene rendering.
\newblock \emph{arXiv preprint arXiv:2310.08528}, 2023.

\bibitem[Xu et~al.(2020)Xu, Bazavan, Zanfir, Freeman, Sukthankar, and
  Sminchisescu]{Xu_2020_CVPR}
Hongyi Xu, Eduard~Gabriel Bazavan, Andrei Zanfir, William~T. Freeman, Rahul
  Sukthankar, and Cristian Sminchisescu.
\newblock Ghum \& ghuml: Generative 3d human shape and articulated pose models.
\newblock In \emph{Proc. of CVPR}, 2020.

\bibitem[Xu et~al.(2021)Xu, Alldieck, and Sminchisescu]{HNeRF:NeurIPS:2021}
Hongyi Xu, Thiemo Alldieck, and Cristian Sminchisescu.
\newblock H-ne{RF}: Neural radiance fields for rendering and temporal
  reconstruction of humans in motion.
\newblock In \emph{Proc. of NeurIPS}, 2021.

\bibitem[Xu et~al.(2022)Xu, Xu, Philip, Bi, Shu, Sunkavalli, and
  Neumann]{xu2022point}
Qiangeng Xu, Zexiang Xu, Julien Philip, Sai Bi, Zhixin Shu, Kalyan Sunkavalli,
  and Ulrich Neumann.
\newblock Point-nerf: Point-based neural radiance fields.
\newblock In \emph{Proc. of CVPR}, 2022.

\bibitem[Yang et~al.(2023{\natexlab{a}})Yang, Gao, Zhou, Jiao, Zhang, and
  Jin]{yang2023deformable3dgs}
Ziyi Yang, Xinyu Gao, Wen Zhou, Shaohui Jiao, Yuqing Zhang, and Xiaogang Jin.
\newblock Deformable 3d gaussians for high-fidelity monocular dynamic scene
  reconstruction.
\newblock \emph{arXiv preprint arXiv:2309.13101}, 2023{\natexlab{a}}.

\bibitem[Yang et~al.(2023{\natexlab{b}})Yang, Yang, Pan, Zhu, and
  Zhang]{yang2023gs4d}
Zeyu Yang, Hongye Yang, Zijie Pan, Xiatian Zhu, and Li Zhang.
\newblock Real-time photorealistic dynamic scene representation and rendering
  with 4d gaussian splatting.
\newblock \emph{arXiv preprint arXiv 2310.10642}, 2023{\natexlab{b}}.

\bibitem[Yariv et~al.(2020)Yariv, Kasten, Moran, Galun, Atzmon, Ronen, and
  Lipman]{yariv2020multiview}
Lior Yariv, Yoni Kasten, Dror Moran, Meirav Galun, Matan Atzmon, Basri Ronen,
  and Yaron Lipman.
\newblock Multiview neural surface reconstruction by disentangling geometry and
  appearance.
\newblock In \emph{Proc. of NeurIPS}, 2020.

\bibitem[Yariv et~al.(2021)Yariv, Gu, Kasten, and Lipman]{volsdf:NeurIPS:2021}
Lior Yariv, Jiatao Gu, Yoni Kasten, and Yaron Lipman.
\newblock Volume rendering of neural implicit surfaces.
\newblock In \emph{Proc. of NeurIPS}, 2021.

\bibitem[Yariv et~al.(2023)Yariv, Hedman, Reiser, Verbin, Srinivasan, Szeliski,
  Barron, and Mildenhall]{yariv2023bakedsdf}
Lior Yariv, Peter Hedman, Christian Reiser, Dor Verbin, Pratul~P. Srinivasan,
  Richard Szeliski, Jonathan~T. Barron, and Ben Mildenhall.
\newblock Bakedsdf: Meshing neural sdfs for real-time view synthesis.
\newblock In \emph{Proc. of SIGGRAPH}, 2023.

\bibitem[Ye et~al.(2023)Ye, Shao, and Zhou]{ye2023animatable}
Keyang Ye, Tianjia Shao, and Kun Zhou.
\newblock Animatable 3d gaussians for high-fidelity synthesis of human motions,
  2023.

\bibitem[Yu et~al.(2021)Yu, Li, Tancik, Li, Ng, and
  Kanazawa]{yu2021plenoctrees}
Alex Yu, Ruilong Li, Matthew Tancik, Hao Li, Ren Ng, and Angjoo Kanazawa.
\newblock {PlenOctrees} for real-time rendering of neural radiance fields.
\newblock In \emph{Proc. of ICCV}, 2021.

\bibitem[Yu et~al.(2023)Yu, Cheng, Liu, Wu, and Lin]{yu2023monohuman}
Zhengming Yu, Wei Cheng, xian Liu, Wayne Wu, and Kwan-Yee Lin.
\newblock {MonoHuman}: Animatable human neural field from monocular video.
\newblock In \emph{Proc. of CVPR}, 2023.

\bibitem[Zhang et~al.(2022)Zhang, Baek, Rusinkiewicz, and
  Heide]{zhang2022differentiable}
Qiang Zhang, Seung-Hwan Baek, Szymon Rusinkiewicz, and Felix Heide.
\newblock Differentiable point-based radiance fields for efficient view
  synthesis.
\newblock In \emph{SIGGRAPH Asia Conference Proceedings}, 2022.

\bibitem[Zhao et~al.(2022)Zhao, Jiang, Yao, Zhang, Wang, Dai, Zhong, Zhang, Wu,
  Xu, and Yu]{NSR:SIGASIA:2022}
Fuqiang Zhao, Yuheng Jiang, Kaixin Yao, Jiakai Zhang, Liao Wang, Haizhao Dai,
  Yuhui Zhong, Yingliang Zhang, Minye Wu, Lan Xu, and Jingyi Yu.
\newblock Human performance modeling and rendering via neural animated mesh.
\newblock \emph{ACM Transactions on Graphics, (Proc. SIGGRAPH Asia)},
  41\penalty0 (6), 2022.

\bibitem[Zheng et~al.(2023)Zheng, Yifan, Wetzstein, Black, and
  Hilliges]{Zheng2023pointavatar}
Yufeng Zheng, Wang Yifan, Gordon Wetzstein, Michael~J. Black, and Otmar
  Hilliges.
\newblock Pointavatar: Deformable point-based head avatars from videos.
\newblock In \emph{Proc. of ECCV}, 2023.

\bibitem[Zheng et~al.(2022)Zheng, Huang, Yu, Zhang, Guo, and
  Liu]{zheng2022structured}
Zerong Zheng, Han Huang, Tao Yu, Hongwen Zhang, Yandong Guo, and Yebin Liu.
\newblock Structured local radiance fields for human avatar modeling.
\newblock In \emph{Proc. of CVPR}, 2022.

\bibitem[Zielonka et~al.(2023)Zielonka, Bagautdinov, Saito, Zollhöfer, Thies,
  and Romero]{Zielonka2023Drivable3D}
Wojciech Zielonka, Timur Bagautdinov, Shunsuke Saito, Michael Zollhöfer,
  Justus Thies, and Javier Romero.
\newblock Drivable 3d gaussian avatars.
\newblock \emph{arXiv preprint arXiv:2311.08581}, 2023.

\bibitem[Zwicker et~al.(2001)Zwicker, Pfister, van Baar, and
  Gross]{zwicker2001ewa}
M. Zwicker, H. Pfister, J. van Baar, and M. Gross.
\newblock Ewa volume splatting.
\newblock In \emph{Proceedings Visualization, 2001. VIS '01.}, pages 29--538,
  2001.

\end{thebibliography}
